\RequirePackage{fix-cm}
\documentclass[twocolumn]{svjour3}          
\smartqed  
\usepackage{graphicx}
\usepackage{amssymb,amsfonts}%
\usepackage[authoryear]{natbib}
\usepackage{xcolor}
\definecolor{citeblue}{RGB}{0,102,204}
\usepackage[colorlinks=true,linkcolor=citeblue,citecolor=citeblue,urlcolor=citeblue]{hyperref}

\usepackage{amsmath}
\usepackage{graphicx}%
\usepackage{multirow}%
\usepackage{mathrsfs}%
\usepackage[title]{appendix}%
\usepackage{textcomp}%
\usepackage{arydshln}
\usepackage{manyfoot}%
\usepackage{booktabs}%
\usepackage{algorithm}%
\usepackage{algorithmicx}%
\usepackage{listings}%
\usepackage{microtype}
\usepackage{amsthm}
\usepackage{amssymb}
\bibpunct{(}{)}{;}{a}{,}{,}
\let\cite\citep

\newcommand{\trans}{\mathsf{T}}

\makeatletter
\let\SVJorig@maketitle\@maketitle
\def\@maketitle{%
  \SVJorig@maketitle
  \vspace{-0.2em}
  \begin{center}
    \includegraphics[width=\textwidth,height=.7\textheight,keepaspectratio]{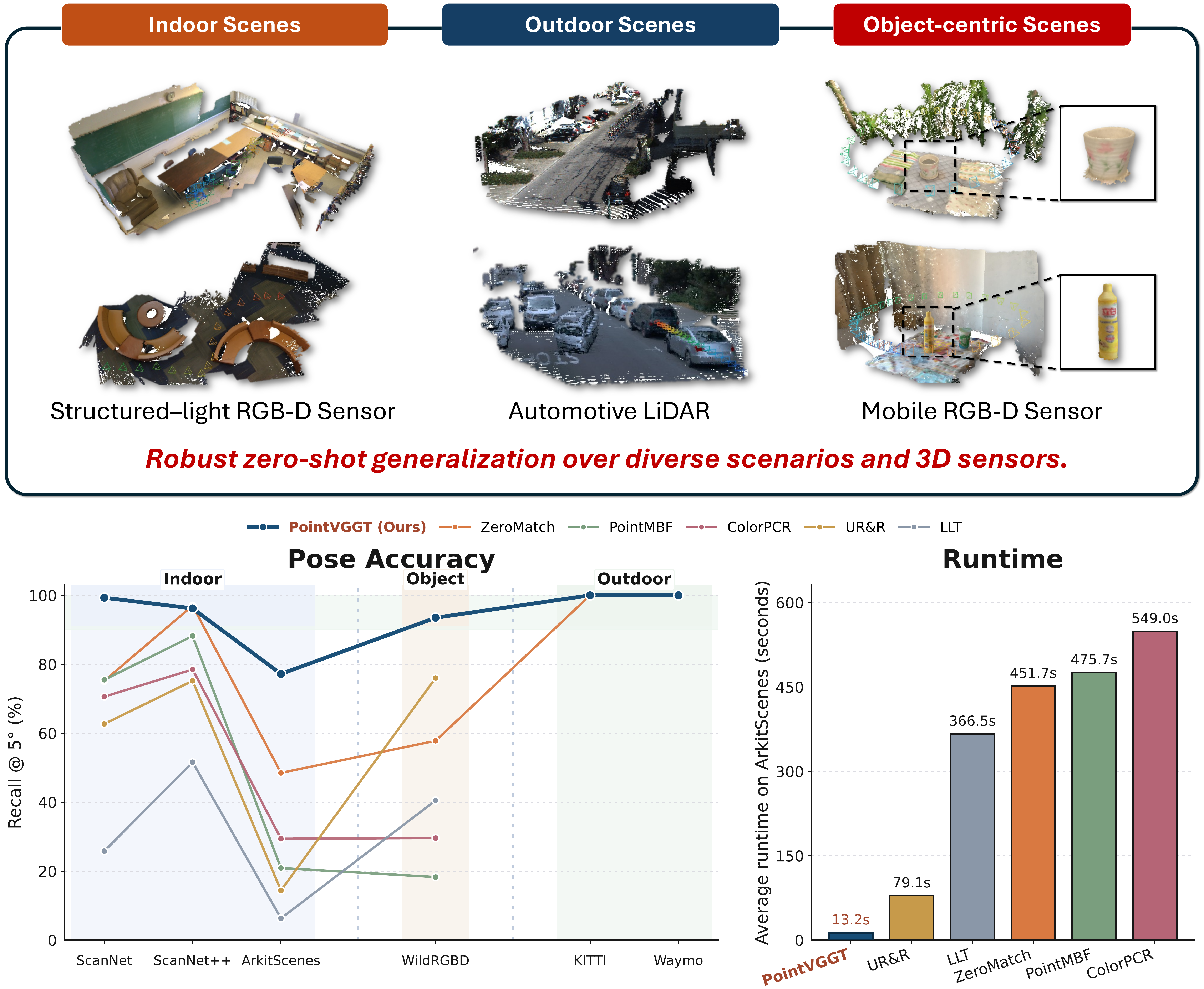}
  \end{center}
  \vspace{0.2em}
  \refstepcounter{figure}\label{fig:teaser}
  \vspace{-4mm}
  {\small
  \noindent\textbf{\figurename~\thefigure.}
PointVGGT enables robust and efficient zero-shot multiview registration across diverse scenarios and sensor types, 
consistently achieving superior or competitive pose accuracy compared to prior methods across indoor, outdoor, and object-centric benchmarks while running substantially faster.
\par}
  \vspace{-0.2em}
}
\makeatother

\begin{document}
\title{PointVGGT: Zero-Shot Multiview RGB-D Point Cloud Registration with Visual Geometry Foundation Priors}


\author{Haobo Jiang \and Liang Yu \and Jianmin Zheng
}


 \institute{
 	        Haobo Jiang \at
            Nanyang Technological University, Singapore \\
            \email{haobo.jiang@ntu.edu.sg}           
            \and
            Liang Yu  \at
            Alibaba Group, China \\
            \email{liangyu.yl@alibaba-inc.com}
            \and
           Jianmin Zheng  \at
           Nanyang Technological University, Singapore \\
           \email{ASJMZheng@ntu.edu.sg}
 }

\date{Received: date / Accepted: date}

\maketitle


\begin{abstract}
This paper addresses multiview RGB-D point cloud registration, aiming to estimate global rigid poses for unordered RGB-D scans and align them in a metrically consistent coordinate frame. The conventional \textit{pairwise-then-global} paradigm suffers from locally optimized pairwise registration, severe error propagation and high computational burden. In particular, existing methods typically treat RGB data as a mere auxiliary matching cue and overlook the holistic geometric priors (e.g., camera poses and 3D models) encoded across image sequences. 
This paper introduces PointVGGT, a zero-shot framework built upon a novel \emph{foundation-then-refinement} paradigm that systematically leverages visual geometry foundation models (e.g., VGGT) as the computational backbone for robust, training-free multiview RGB-D registration. 
In the \textit{foundation} stage, we directly recover metrically consistent global poses (without any pairwise estimation) by grounding the scale-ambiguous pose predictions of the foundation model against metric depth observations. 
In the  \textit{refinement} stage, we introduce an efficient voxelized spatial hashing mechanism that exploits the globally coherent 3D reconstruction (induced by the foundation model) as a shared spatial anchor, enabling dense multiview correspondences in near-linear time. 
On top of this, an IRLS-based robust motion-only bundle adjustment is performed using a conjugate gradient solver to jointly minimize the correspondence and reprojection residuals for multiview pose refinement. 
Extensive experiments on indoor/object-centric/outdoor datasets verify the outstanding zero-shot  registration accuracy and computational efficiency of our proposed method. [\href{https://github.com/Jiang-HB/PointVGGT}{Code}]
\end{abstract}



 \begin{figure}[t]
	\centering
	\includegraphics[width=\columnwidth]{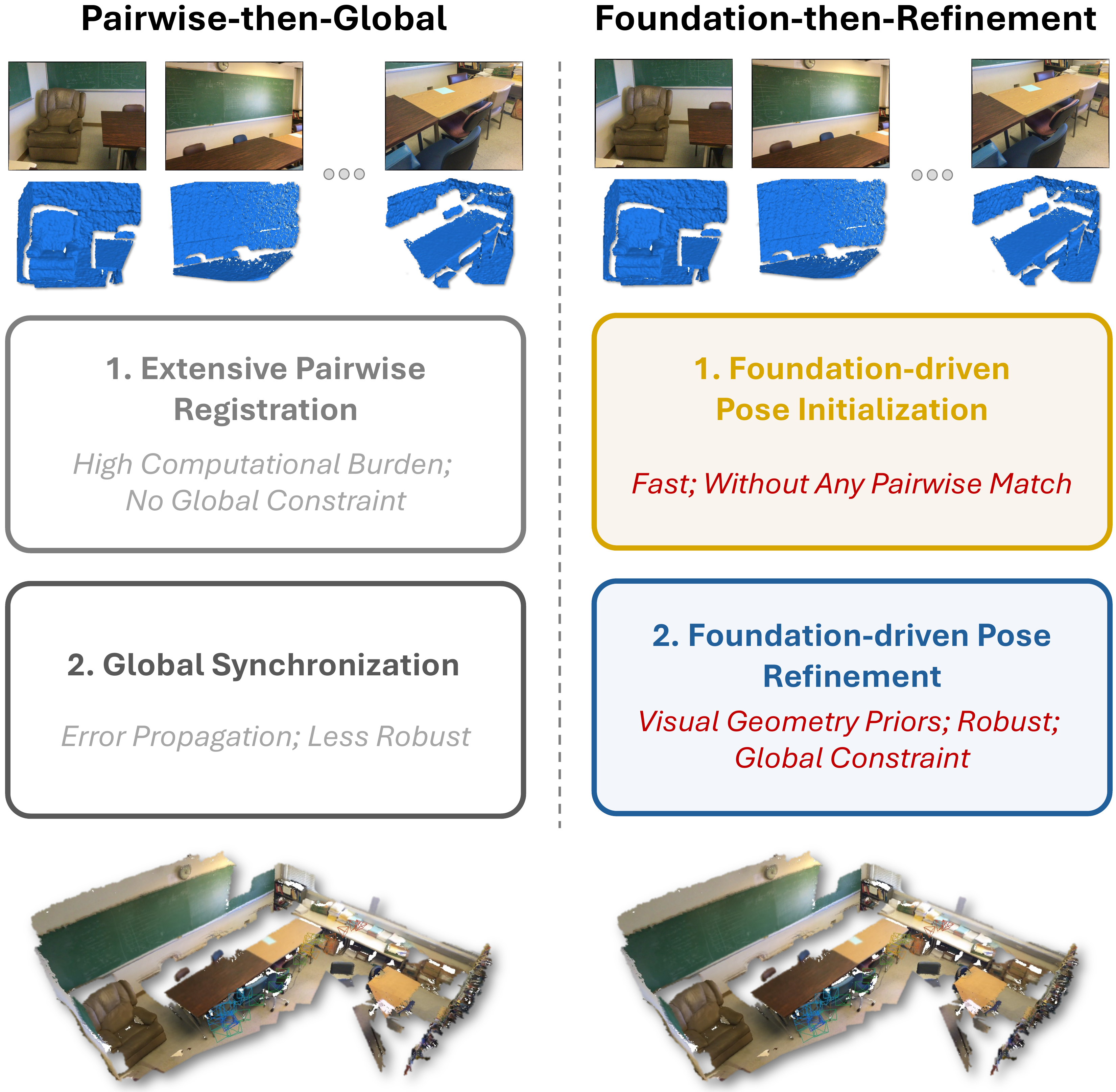}
    \vspace{-4mm}
	\caption{\textbf{Paradigm comparison.} 
    Conventional \textit{pairwise-then-global} pipelines rely on extensive pairwise matching (high computational cost and no global constraint) followed by pose synchronization (error propagation). By contrast, our \textit{foundation-then-refinement} pipeline directly initializes poses from visual geometry foundation priors without any costly pairwise matching, and subsequently refines them with global geometric constraints, achieving a more efficient and robust registration process.}
	\label{motiv}
	\vspace{-5mm}
\end{figure}

\section{Introduction}
\label{intro}
Multiview RGB-D point cloud registration estimates the rigid poses of a set of partially overlapping RGB-D scans and align them into a unified coordinate frame, with extensive applications such as 3D scene reconstruction~\cite{dai2017bundlefusion,newcombe2011kinectfusion,slavcheva2018sdf}, AR/VR~\cite{azuma1997survey,kim2022benchmark}, and embodied AI~\cite{savva2019habitat,szot2021habitat}. However, due to real-world challenges such as low overlap, repetitive structures, and noisy measurements, multiview registration remains both unreliable and computationally expensive, limiting its practical deployment.

Existing multiview registration methods predominantly follow a \emph{pairwise-then-global} paradigm~\cite{arrigoni2016spectral,choi2015robust,govindu2004lie,huang2017translation,wang2023robust}, in which relative transformations are first estimated through extensive pairwise registration across scan pairs~\cite{huang2021predator,qin2022geometric}, and then integrated into globally consistent poses via pose graph synchronization~\cite{wang2013exact,lee2022hara}. Despite its effectiveness, this paradigm suffers from several fundamental limitations, including local optima and high computational cost in pairwise registration, as well as severe error propagation during global synchronization. Moreover, many learning-based methods exhibit limited domain generalization and often require model finetuning to adapt to new environments, further reducing their practicality in real-world applications.

More importantly, within this paradigm, RGB images are typically treated merely as auxiliary cues for improving pairwise feature matching~\cite{yuan2023pointmbf,Jiang_2025_CVPR} and are discarded thereafter. However, recent visual geometry foundation models, such as the pioneering VGGT~\cite{wang2025vggt}, have shown that multiview RGB images inherently encode reliable pose and structural priors by jointly predicting high-quality dense 3D pointmaps and camera poses in a single feed-forward pass, while exhibiting impressive zero-shot generalization. This naturally raises a fundamental question: \emph{Can such rich visual geometry priors be  leveraged for zero-shot multiview RGB-D registration without any domain-specific weight finetuning?}

Motivated by these insights, this paper introduces \textbf{PointVGGT}, a training-free \emph{foundation-then-refinement} framework that bridges the visual geometry priors (encoded in RGB images by VGGT-style foundation models) with multiview RGB-D registration, enabling efficient and robust zero-shot alignment (see Fig.~\ref{motiv}).
Specifically, in the \emph{foundation} stage, we initialize global poses by converting the scale-ambiguous pose predictions of the foundation model into metrically grounded poses through depth-based scale alignment with sensor observations.
In the \emph{refinement} stage, our key observation is that the globally coherent 3D model reconstructed by the foundation model naturally provides rich correspondence priors: geometrically corresponding points from different views tend to cluster tightly in the reconstructed space. Exploiting this property, we develop a voxelized spatial hashing scheme that discretizes the global space into uniform voxel grids and performs GPU-accelerated intra-voxel matching for fast outlier removal and cross-view correspondence building.
Given these correspondences and initial poses, we jointly refine poses via a motion-only bundle adjustment~\cite{triggs1999bundle}, optimized using Geman-McClure kernel-based iterative reweighted least squares (IRLS)~\cite{holland1977robust} and a conjugate gradient solver, to minimize both correspondence and geometric reprojection residuals.

Notably, unlike the conventional \emph{pairwise-then-global} paradigm, our \textit{foundation-then-refinement} pipeline completely eliminates locally biased and computationally intensive pairwise registration, while fully exploiting the holistic pose and geometric priors encoded in RGB inputs for reliable multiview alignment. As a result, PointVGGT reduces inference time from minutes to seconds while maintaining strong robustness. Extensive experiments show that our method achieves outstanding registration accuracy and efficiency in a fully zero-shot setting, outperforming both classical and learning-based approaches without any task-specific training or fine-tuning. 
In summary, our  contributions are as follows: 
\begin{itemize}
	\item We propose a novel and powerful \emph{foundation-then-refinement} paradigm for multiview RGB-D registration that, for the first time, systematically leverages foundation model-driven visual geometry priors encoded in RGB images to enable training-free, zero-shot, and robust multiview registration.
	\vspace{1mm}
	
	\item We develop an efficient foundation-driven pose initialization that grounds the scale-ambiguous pose predictions of foundation models with sensor depth observations, yielding metrically consistent global poses without redundant pairwise registration.
	\vspace{1mm}
	
	\item We design a voxelized spatial hashing scheme that exploits the globally consistent 3D reconstruction induced by the foundation model to efficiently construct dense multiview correspondences in near-linear time,  avoiding time-consuming all-pair matching.
	\vspace{1mm}
	
	\item We introduce a robust motion-only bundle adjustment with Geman-McClure robust kernel-based IRLS and a conjugate gradient solver for efficient and reliable multiview pose refinement.
\end{itemize}
Our proposed \emph{foundation-then-refinement} paradigm is general and inherently plug-and-play. It can be seamlessly integrated with diverse visual VGGT-style 3D foundation models, such as VGGT~\cite{wang2025vggt}, Pi3~\cite{wang2025pi}, and DepthAnything3~\cite{lin2025depth}, naturally benefiting from their continued advances, which enables sustained performance improvements as these foundation models evolve.

\section{Related Work}
\label{sec:relatedwork}
\noindent\textbf{Pairwise Point Cloud Registration.}
Pairwise point cloud registration is the task of recovering the relative rigid transformation between two 3D scans. 
State-of-the-art solutions predominantly rely on correspondence-driven pipelines, where discriminative geometric representations are first constructed and subsequently matched to estimate the relative pose. 
Early methods design handcrafted descriptors to encode local surface geometry~\cite{johnson1999using,tombari2010unique,salti2014shot,rusu2008aligning,rusu2009fast}. 
Representative examples include USC~\cite{tombari2010unique}, which builds shape-context representations under a local reference frame, SHOT~\cite{salti2014shot}, which aggregates normal-based histograms, and PFH/FPFH~\cite{rusu2008aligning,rusu2009fast}, which capture pairwise geometric relationships within local neighborhoods. 

With the development of deep learning, more recent approaches replace handcrafted features with learned representations derived from large-scale data~\cite{jiang2026gm,wang2023zero,li2022lepard,xu2026robust,choy2020deep,chen2023sira,jiang2023se,jiang2023robust,jiang2021sampling}. 
For instance, FCGF~\cite{choy2019fully} learns fully-convolutional sparse 3D features for reliable matching, while D3Feat~\cite{bai2020d3feat} jointly learns keypoint detection and description to improve matching quality.
CoFiNet~\cite{yu2021cofinet} adopts a coarse-to-fine matching strategy to refine correspondences progressively, and Predator~\cite{huang2021predator} leverages overlap-aware cross-attention to improve matching robustness in low-overlap situations.
RoITr~\cite{yu2023rotation} enhances robustness via PPF-based rotation-invariant feature encoding, and GeoTrans~\cite{qin2022geometric} integrates explicit geometric embeddings for more discriminative correspondence reasoning. 
PARE-Net~\cite{yao2024pare} develops position-aware rotation-equivariant backbone for efficient and robust matching.       
Gen-PCR~\cite{jiang2025generative} designs Match-ControlNet to generate free-lunch RGB cues for semantics-enhanced matching. 

\noindent\textbf{Pairwise RGB-D Point Cloud Registration.} Existing pairwise RGB-D registration approaches can be broadly grouped into the optimization-based methods and learning-based methods. {Optimization-based approaches} predominantly extend the ICP framework~\cite{besl1992method} by augmenting geometric alignment with color-derived constraints. One line of work modifies the objective function itself: Park et al.~\cite{park2017colored} and Whelan et al.~\cite{whelan2015real} append a photometric residual to the standard point-to-plane cost, while Danelljan et al.~\cite{danelljan2016probabilistic} embed color within a Gaussian mixture model for probabilistic EM-based optimization, and Semantic-ICP~\cite{parkison2018semantic} treats semantic labels as latent variables in an EM-ICP formulation~\cite{granger2002multi}. A complementary line of work instead enriches the correspondence stage: Korn et al.~\cite{korn2014color} concatenate color channels with spatial coordinates to form augmented descriptors, Godin et al.~\cite{godin2001method} construct transformation-invariant photometric attributes to constrain nearest-neighbor selection, Servos et al.~\cite{servos2014multi} exploit both color and intensity for alignment, and GICP-RKHS~\cite{parkison2019boosting} enforces intensity-surface consistency through reproducing-kernel Hilbert space regularizers. 

  {Learning-based approaches} replace handcrafted fusion
  strategies with data-driven feature representations. UR\&R~\cite{el2021unsupervisedr} learns transformations through
  differentiable alignment and rendering, enforcing
  geometric and photometric consistency without explicit pose
  supervision. BYOC~\cite{el2021bootstrap} bootstraps feature
  learning by using visual correspondences as self-generated
  supervisory signals. Subsequent methods focus on more effective
  cross-modal fusion: LLT~\cite{wang2022improving} progressively merges color and geometry by employing multi-scale local linear transformations;   PointMBF~\cite{yuan2023pointmbf} introduces
  bidirectional multi-scale fusion to capture complementary
  information across the two modalities;
  ColorPCR~\cite{mu2024colorpcr} applies multi-stage
  geometric-color integration to enhance matching robustness. An alternative
  direction is explored by NeRF-UR~\cite{yu2024nerf}, which adopts a
  neural radiance field as a global scene model for unsupervised
  frame-to-model training. ZeroMatch~\cite{Jiang_2025_CVPR} exploits pretrained Stable Diffusion to extract transferable zero-shot visual representations for general RGB-D matching. 
  \cite{shou2025unsupervised} introduces the overlapping constraint for inliers detection, improving robustness. 
Beyond pairwise RGB-D registration, our multiview setting requires jointly reasoning about geometric consistency across all views, substantially increasing both problem difficulty and computational complexity. 
Rather than merely treating RGB  as the auxiliary matching cue in prior RGB-D methods, we explicitly exploit its inherent visual geometry priors for improved  multiview matching.

 \begin{figure*}[t]
	\centering
	\includegraphics[width=\textwidth]{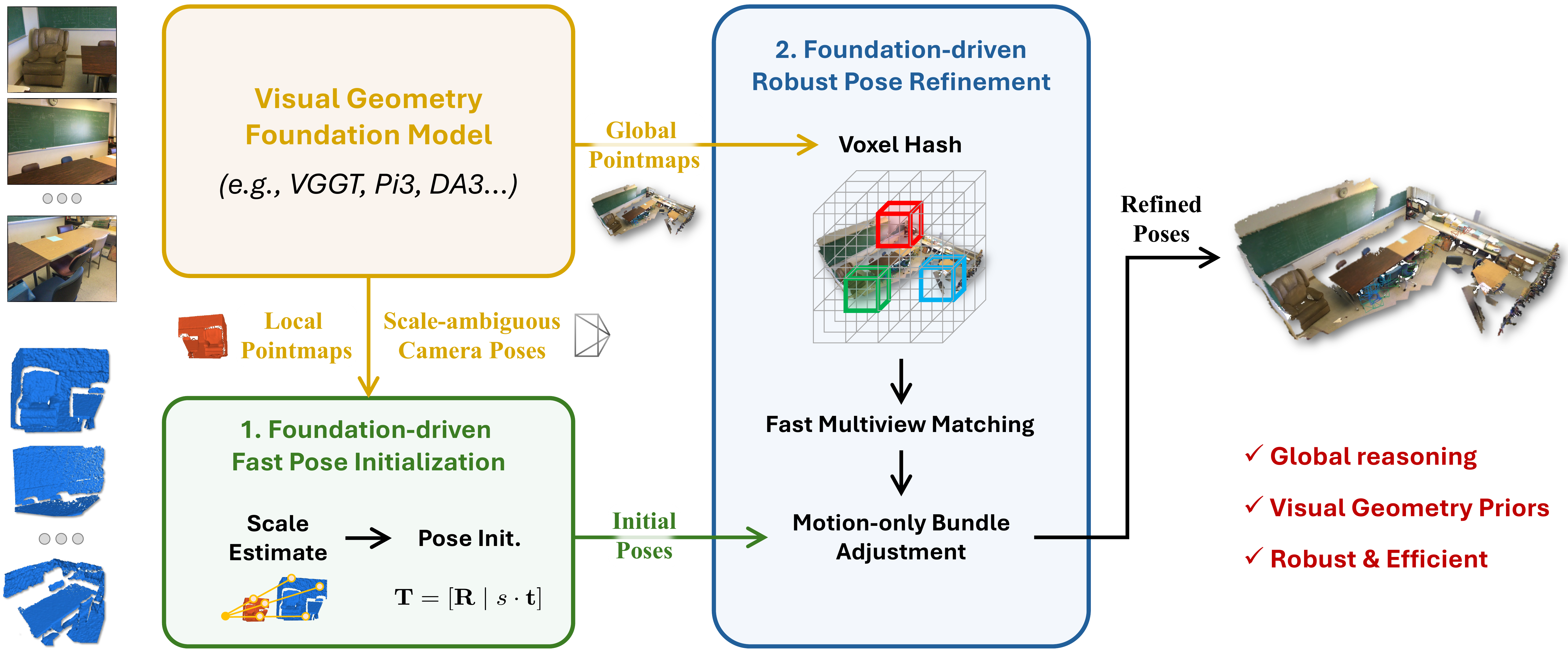}
	\vspace{-3mm}
	\caption{\textbf{Main Pipeline of PointVGGT}. Given a collection of unordered RGB-D scans, PointVGGT first processes the RGB sequence through a visual geometry foundation model to obtain scale-ambiguous camera poses and local/global 3D pointmaps. Then, foundation-driven pose initialization module grounds them into metric space via depth-based scale estimation for pose initialization. In the foundation-driven robust pose refinement module, the globally coherent reconstruction is exploited by a voxelized spatial hashing scheme to construct dense cross-view correspondences, which are then refined through an IRLS-based motion-only bundle adjustment. Our design eliminates pairwise matching, enabling zero-shot registration (without domain-specific weight finetuning) with outstanding accuracy and efficiency.}
	\label{pipeline}
	\vspace{-3mm}
\end{figure*}

\noindent\textbf{Multiview Point Cloud Registration.}    
  The multiview registration pipelines primarily follow a \textit{pairwise-then-global} paradigm: pairwise relative pose estimation followed by global pose synchronization. 
  Early methods~\cite{huber2003fully,choi2015robust} construct a
  fully connected pose graph from pairwise transformations, then
  recover global poses by solving a synchronization problem over
  $\mathrm{SE}(3)$. Because pairwise estimates are inevitably
  corrupted by matching failures, robust synchronization has received
  sustained attention. One direction isolates rotation averaging:
  \cite{govindu2004lie} proposes Lie-algebraic averaging,
  \cite{hartley2011l1} introduces an L1 formulation,
  and \cite{wang2013exact} provides convex relaxations with recovery guarantees, with translation solved
  subsequently~\cite{huang2017translation}. Joint $\mathrm{SE}(3)$
  synchronization has also been pursued through non-linear
  optimization~\cite{kummerle2011g,grisetti2011tutorial}, Bayesian
  inference~\cite{birdal2018bayesian}, spectral
  methods~\cite{arrigoni2016spectral}, and hierarchical
  strategies~\cite{lee2022hara}. Despite their theoretical elegance,
  these methods inherit a fundamental fragility: synchronization
  quality is upper-bounded by the pairwise estimates it receives, and
  erroneous edges propagate and amplify through the graph.

  Recent methods integrate learned components to mitigate this
  fragility. Gojcic et al.~\cite{gojcic2020learning} jointly learn
  descriptors and a differentiable synchronization module. Huang et
  al.~\cite{huang2019learning} and Yew and
  Lee~\cite{yew2021learning} learn to reweight edges during
  synchronization. SGHR~\cite{wang2023robust} predicts overlap ratios
  for pose graph initialization with history-based reweighting.
  FeatSync~\cite{hu2024featsync} refines synchronized poses through
  attention-based feature alignment.
  MDGD~\cite{li2024matching} fuses descriptor and geometric cues for
  reliable graph construction, and Jin et
  al.~\cite{jin2024multiway} formulate multiway registration via
  diffusion and global optimization. While these methods improve
  robustness, they remain anchored to the
  pairwise-then-global architecture, inheriting both its
  quadratic cost and susceptibility to cascaded errors. 
While FUSER~\cite{jiang2025fuser} introduces a feed-forward paradigm to alleviate this issue, it is limited to geometry-only registration and cannot generalize to the RGB-D setting. 
  Our approach departs from this paradigm entirely: globally coherent
  poses are obtained in a single feed-forward pass via a visual
  geometry foundation model, then refined through holistic multiview
  optimization, eliminating the time-consuming pose graph establishment and the error propagation during pose synchronization. 

  \noindent\textbf{Visual Geometry Foundation Models.}
Visual geometry foundation models infer camera geometry and 3D scene structure from an unordered image set in an end-to-end, feed-forward manner.
DUSt3R~\cite{wang2024dust3r} pioneers this direction by regressing dense point maps from image pairs without requiring calibrated cameras or known poses.
Fast3R~\cite{yang2025fast3r} extends this pairwise paradigm to scalable many-view reconstruction by processing multiple images jointly in a single forward pass.
VGGT~\cite{wang2025vggt} further advances this family by unifying the prediction of multiple scene-level geometric attributes, including camera parameters, depth maps, point maps, and 3D point tracks. 
Pi3~\cite{wang2025pi} introduces a perturbation-equivariant architecture that eliminates the common assumption of treating the first frame as the world coordinate frame. 
MapAnything~\cite{keetha2025mapanything} generalizes the input-output interface by supporting optional geometric priors, such as intrinsics, poses, and depth, and by predicting a fully factored metric representation of scene geometry.
Relatedly, DepthAnything3 (DA3)~\cite{lin2025depth} adopts a more minimalist design, using a plain Transformer and a unified depth-ray representation to recover spatially consistent geometry from different visual inputs. More advanced VGGT variants have been explored in~\cite{shen2025fastvggt,lu2026ovggt,wang2025flashvggt,yuan2026infinitevggt}. In this paper, we aim to integrate the visual geometry priors predicted by such foundation models for robust and efficient multiview RGB-D point cloud registration.

\section{Methodology}
\subsection{Problem Statement}
\noindent\textbf{Problem Definition.}
Given a set of unordered, partially overlapping RGB-D scans  $\mathcal{V}=\{(\mathbf{I}_i, \mathbf{X}_i)\}_{i=1}^{S}$, where $\mathbf{I}_i\in\mathbb{R}^{H_{{rgb}}\times W_{{rgb}}\times 3}$ denotes the RGB image and $\mathbf{X}_i\in\mathbb{R}^{N_{i}\times 3}$ represents the point cloud (back-projected from the metric depth map $\mathbf{D}_i\in\mathbb{R}^{H_{dpt}\times W_{dpt}}$), the multiview RGB-D point cloud registration seeks to estimate globally consistent rigid poses $\mathcal{T}=\{T_i=[\mathbf{R}_i\mid \mathbf{t}_i]\in{SE}(3)\}_{i=1}^{S}$ that align all scans into a common world coordinate frame. Here, $\mathbf{R}_i\in{SO}(3)$ and $\mathbf{t}_i\in\mathbb{R}^3$ represent the 3D rotation matrix and 3D translation vector for each view $i$, respectively. 

\noindent\textbf{Multiview Registration Pipeline.}  
Most existing  multiview registration approaches~\cite{choi2015robust,huang2017translation,wang2023robust} decompose this problem into two sequential stages. The first stage estimates  relative poses $\hat{\mathbf{T}}_{i\leftarrow j}$ between paired views $(i, j)$ via  feature matching followed by robust estimators (\emph{e.g.}, RANSAC). 
The second stage then organizes these pairwise pose estimates as edges $\mathcal{E}$ into a pose graph $\mathcal{G}=(\mathcal{V},\mathcal{E})$, and performs  transformation synchronizes to recover per-scan global poses. 
Notably, existing methods are largely geometry-centric, where the RGB-D setting is typically treated as a mere modality enhancement for improved pairwise matching~\cite{yuan2023pointmbf,Jiang_2025_CVPR} rather than as an independent research problem. 

\noindent\textbf{Visual Geometry Foundation Priors.}  
Given the unordered RGB images $\{\mathbf{I}_i\}_{i=1}^S$, recent 3D foundation models (e.g., VGGT~\cite{wang2025vggt}) recover visual geometry priors in a feed-forward manner, including per-scan global camera poses $\{\hat{\mathbf{T}}^{FM}_i=[\hat{\mathbf{R}}^{FM}_i|\hat{\mathbf{t}}^{FM}_i]\in SE(3)\}_{i=1}^S$, local pointmaps $\{{\mathbf{L}}_i\in\mathbb{R}^{H\times W\times 3}\}_{i=1}^S$, and global pointmaps $\{{\mathbf{G}}_i\in\mathbb{R}^{H\times W\times 3}\}_{i=1}^S$. 
Here, a pointmap defines a dense one-to-one mapping between image pixels and 3D points, while ``local'' and ``global'' denote the camera and world coordinate systems, respectively. 
Importantly, for RGB-only reconstruction, the recovered translations $\{\hat{\mathbf{t}}^{FM}_i\}$ and the overall scene geometry (like pointmaps) are generally defined only up to an unknown global scale, since pure RGB observations do not provide absolute metric depth cues. 
Recent extensions, \emph{e.g.} MapAnything~\cite{keetha2025mapanything} and DepthAnything3~\cite{lin2025depth}, further improve metric geometric recovery by leveraging richer additional geometric inputs (\emph{e.g.}, depth maps). Nevertheless, in unconstrained real-world settings, feed-forward reconstructions may still suffer from residual scale ambiguity or cross-view scale inconsistency.

\subsection{Motivation}\label{motivate}
Despite the importance of RGB-D sensing for robust registration~\cite{steinbrucker2011real,dai2017bundlefusion}, multiview RGB-D point cloud registration remains surprisingly underexplored. Rather than being treated as an independent research problem, it is often regarded as a straightforward extension of geometry-only multiview registration  that follows the conventional \textit{pairwise-then-global} paradigm. Within this framework, RGB information is typically incorporated merely as an auxiliary signal to enhance feature matching for improved pairwise registration, rather than being jointly modeled as a core component of the multiview registration process.

While effective to some extent, this paradigm inherently suffers from a series of fundamental limitations:
\noindent\textbf{(i) Suboptimal exploitation of visual geometry:} Existing methods typically restrict RGB data to only feature-level fusion for improved pairwise matching. Such an operation discards the rich multiview geometric priors (\emph{e.g.}, camera poses and 3D models) embedded across image sequences, as evidenced by  SfM pipelines~\cite{schonberger2016structure} and recent visual foundation models~\cite{wang2025pi}, which are crucial for global pose reasoning. 
Confining them to only local feature  matching wastes such structural potential and truncates the upper bound of registration precision;   
\noindent\textbf{(ii) Local optima and global error propagation:}  
Pairwise alignment optimizes each scan pair in isolation, lacking any mechanism to incorporate global geometric constraints from other views. Consequently, in geometrically ambiguous settings, such as cases with low overlap or repetitive patterns, correspondence ambiguities cannot be resolved through joint cross-view geometric reasoning, rendering the pairwise optimization susceptible to local minima.  Furthermore, these locally biased pose estimations then enter the global synchronization stage, where pose error is propagated and accumulated across the pose graph, ultimately resulting in severely degraded global pose estimations;   
\noindent\textbf{(iii) Severe computational overhead:}  
The existing paradigm depends on constructing a dense and reliable pose graph through numerous pairwise registrations. Each involves costly operations such as feature extraction, matching, and outlier removal. As the number of scans increases, these costs grow combinatorially, rendering the entire pipeline inherently incompatible with large-scale or time-sensitive applications. 


To address these challenges, we propose PointVGGT, a novel and powerful framework for zero-shot multiview RGB-D point cloud registration. Unlike the conventional \textit{pairwise-then-global} paradigm, PointVGGT follows a new \textit{foundation-then-refinement} strategy. Specifically, in the \emph{foundation} stage (Sec.~\ref{init}), instead of treating RGB images as mere local matching cues, we leverage visual geometry foundation models to extract reliable multiview pose priors encoded across RGB sequences, enabling fast and robust coarse global pose initialization, thereby mitigating limitation \textbf{(i)}. In the \emph{refinement} stage  (Sec.~\ref{ba}), building upon this initialization, we further develop a foundation-driven voxelized spatial hashing mechanism for efficient cross-view correspondence construction, coupled with a motion-only bundle adjustment to jointly refine all scan poses. This design avoids time-consuming pairwise registration and prevents error propagation during  pose synchronization,  addressing limitations \textbf{(ii)} and \textbf{(iii)}. The overall pipeline is illustrated in Fig.~\ref{pipeline}. 

\subsection{Foundation-Driven Fast  Pose Initialization}\label{init}
Reliable camera pose initialization serves as the critical starting point of our \emph{foundation-then-refinement} paradigm.
Inspired by the remarkable pose accuracy achieved by recent visual geometry foundation models (\emph{e.g.}, VGGT~\cite{wang2025vggt} and Pi3~\cite{wang2025pi}), we introduce such models for multiview registration initialization. Specifically, the model processes the entire RGB sequence in a feed-forward manner and directly recovers a set of geometrically consistent camera poses $\{\hat{\mathbf{T}}^{FM}_i = [\hat{\mathbf{R}}^{FM}_i \mid \hat{\mathbf{t}}^{FM}_i]\}$.   This strategy fully exploits the multiview geometric constraints inherently encoded in RGB images for global registration, rather than treating RGB merely as auxiliary matching cues in traditional \emph{pairwise-then-global} pipelines. 

 Nevertheless, due to the intrinsic scale ambiguity of monocular 2D observations, the predicted poses are determined \emph{only up to an unknown global scale factor}. We thus need to ground these scale-ambiguous predictions into the metric coordinate system of the RGB-D sensor, enabling metrically consistent global pose initialization.

\noindent\textbf{Foundation-Driven Rotation Initialization.} 
Since camera rotations are inherently scale-invariant, the rotations predicted by the foundation model remain metrically valid despite the global scale ambiguity. We therefore directly adopt them as the  rotation initialization:
\begin{equation}
	\hat{\mathbf{R}}_i^{{(0)}} = \hat{\mathbf{R}}_i^{FM}\in SO(3), 
	\quad i \in \{1, \dots, S\}.
\end{equation}

\noindent\textbf{Foundation-Driven Translation Initialization.}  
Although the foundation model predicts poses up to an unknown global scale, both the 
foundation-predicted local 3D points $\{\mathbf{l}_{i, k}\in{\mathbf{L}}_{i}\}$ and the metric depth back-projected points $\{\mathbf{x}_{i, k}\in{\mathbf{X}}_{i}\}$ are expressed in the same camera coordinate frame. Under the assumption of a uniform global scale, they satisfy $\mathbf{x}_{i, k} \approx s \, \mathbf{l}_{i, k}$ for each point index $k$. 
Therefore, the global scale can be estimated by enforcing consistency between their radial magnitudes. Specifically, we estimate a single scale parameter by solving the objective below:
\begin{equation}
	s^\star = \arg\min_s \sum_{i=1}^{S}\sum_{k=1}^{H\times W} \left| \frac{\|\mathbf{x}_{i, k}\|_2}{\|\mathbf{l}_{i, k}\|_2} - s \right|,
\end{equation}
which reduces to a 1-D least-absolute-deviation scale estimation problem, for which an optimal solution is given by any median of the ratios:
\begin{equation}\label{eq:scale}
	\hat{s} = \mathrm{median} \left( \left\{\frac{\|\mathbf{x}_{i, k}\|_2}{\|{\mathbf{l}}_{i, k}\|_2}\right\}_{i\in\{1...S\}, k\in\{1...H\times W\}} \right).
\end{equation} 
This estimator provides robustness against depth noise, prediction errors, and occasional mismatches, yielding a stable global scale estimate without requiring any pairwise alignment. 
With the recovered scale factor, we obtain metrically grounded coarse poses by rescaling the translation component: \begin{equation}
	\hat{\mathbf{t}}_i^{{(0)}} = \hat{s}\cdot\hat{\mathbf{t}}_i^{FM}\in\mathbb{R}^3, 
	\quad i \in \{1, \dots, S\}.
\end{equation} This direct grounding step transforms scale-ambiguous foundation predictions into metrically consistent initial poses $\{\hat{\mathbf{T}}^{(0)}_i=[\hat{\mathbf{R}}_i^{{(0)}}|\hat{\mathbf{t}}_i^{{(0)}}]\}$, bypassing the pairwise registration strategy. 

Notably, this initialization step is entirely registration-free, thereby substantially reducing the computational cost compared with conventional \textit{pairwise-then-global} pipelines. Moreover, thanks to large-scale data-driven training and holistic multiview reasoning of foundation models, the predicted poses exhibit impressive robustness under challenging conditions such as low overlap and repetitive patterns, providing a stable and reliable starting point for subsequent pose refinement.

\subsection{Foundation-Driven Motion-only Bundle Adjustment For Pose Refinement}\label{ba}
Although the proposed foundation-driven fast pose initialization provides reliable coarse pose estimates, its translation component heavily relies on global scale estimation, whose precision would be potentially influenced by real-world sensing noise and point-map inaccuracies, thereby degrading the reconstruction quality. To mitigate this issue, we further develop a foundation-driven bundle adjustment strategy for fine-grained pose refinement.  
Our refinement framework consists of two key components: 
\textbf{(i)} voxelized spatial hashing for fast cross-view correspondence construction, and 
\textbf{(ii)} robust motion-only bundle adjustment, as detailed below.

\subsubsection{Voxelized Spatial Hashing for Fast Cross-view Correspondence Construction}\label{corr}
Establishing dense, reliable cross-view correspondences, that is identifying mutually consistent feature matches across multiple scans, is a prerequisite for subsequent bundle adjustment-based pose refinement. To this end, we introduce a novel and effective voxelized spatial hashing mechanism for efficient, dense and high-quality cross-view correspondence construction. 

\noindent\textbf{Pointmap-as-descriptor.} 
Instead of relying on the computationally intensive extraction of pointwise geometric descriptors per scan for cross-view matching, we directly employ the readily available global pointmaps $\{\mathbf{G}_i\}_{i=1}^S$, inferred by visual foundation models, as 3D descriptors. This is based on the fact that per-view global pointmaps represent the spatial position of each pixel in the world frame. Consequently, correspondences across multiple views naturally exhibit close 3D coordinates, aligning perfectly with the objective of geometric descriptors. Therefore, we utilize the global pointmap as a compact 3D geometric descriptor, denoted as $\{\mathbf{g}_{i,k}\in\mathbb{R}^3\}$, for the per-scan points $\{\mathbf{x}_{i,k}\}$, where their intrinsic pixel alignment ensures overhead-free, one-to-one assignment, and $i\in\{1,...,S\}$ and $k\in\{1, ..., H\times W\}$ represent the scan index and point index, respectively.  

Although our pointmap-as-descriptor strategy provides a compact representation for accelerated matching, conventional pairwise matching strategy across all views still suffers from high computational demands: \textbf{(i)} View-level scalability: Ideally, correspondences should be established for all overlapping scan pairs to ensure sufficient geometric constraints. Given $S$ input views, prior pairwise matching evaluates up to $\mathcal{O}(S^2)$ scan pairs. As $S$ increases, the number of pairwise matching operations grows quadratically, leading to substantial computational overhead; \textbf{(ii)} Point-level scalability: For each scan pair, dense correspondence construction requires a nearest-neighbor search over $N$ points per scan, resulting in $\mathcal{O}(N^2)$ distance evaluations. 
Consequently, the total computational complexity reaches $\mathcal{O}(S^2N^2)$, which quickly becomes prohibitive for large-scale scenes.

%
%



\noindent\textbf{Voxelized Spatial Hashing.}  
To bypass this computational bottleneck, we introduce a voxelized spatial hashing mechanism for fast cross-view correspondence construction. Our core idea relies on discretizing the 3D descriptors (i.e., the global pointmaps) into a uniform voxel grid. Because the global pointmaps are inherently aligned and exhibit strong geometric coherence within a common coordinate frame, potential corresponding points from multiple views naturally cluster within the same voxel. Consequently, we can rapidly identify potential correspondence candidates while filtering out numerous non-corresponding points, thereby significantly improving matching efficiency. Specifically, the 3D descriptor $\mathbf{g}_{i,k}$ of each point $\mathbf{x}_{i,k}$ is assigned to a voxel cell: 
  \begin{equation}\label{voxel}
	\mathcal{V}(\mathbf{g}_{i,k})
	= \left\lfloor \frac{\hat{s}\cdot\mathbf{g}_{i,k}}{\nu} \right\rfloor
\end{equation}
where $\hat{s}$ is the scale factor (estimated in Eq.~\ref{eq:scale}) used to project the global pointmaps into a metric space; $\lfloor\cdot\rfloor$ denotes the floor operation; $\nu$ is the voxel size controlling the spatial granularity; and $\mathcal{V}(\mathbf{g}_{i,k}) \in \mathbb{Z}^3$ denotes the resulting
voxel index. Subsequently, all formed different 3D voxel indices serve as the keys in a spatial hash table ${hash}(\cdot)$, grouping all points sharing the same voxel into a single hash bin, for $\mathcal{O}(1)$ retrieval:
\begin{equation}
{hash}(\mathcal{V}) = \left\{\mathbf{g}_{i,k}\in\mathbb{R}^3 \mid \mathcal{V}(\mathbf{g}_{i, k}) = \mathcal{V} \right\}.
\end{equation}



\noindent\textbf{Neighborhood Voxel-aware Multiview Matching.}
After constructing the spatial hash table, we perform intra-voxel nearest-neighbor searches for correspondence identification. However, rigid voxelization inevitably introduces boundary artifacts: spatially adjacent points may fall into neighboring voxels, causing a strict intra-voxel search to possibly miss the true nearest neighbor. To circumvent this, we expand the search scope to a $3\times3\times3$ voxel neighborhood, strictly eliminating such boundary artifacts. 
Concretely, for a given point $\mathbf{x}_{i,k}$, we first leverage its 3D descriptor $\mathbf{g}_{i,k}$ to compute its localized voxel index $\mathcal{V}(\mathbf{g}_{i,k})$. We then query its $3\times3\times3$ neighboring voxels, denoted as $\mathcal{N}(\mathcal{V}(\mathbf{g}_{i,k}))$, to retrieve a candidate set of 3D descriptors: $\mathcal{N}_{i, k}=\bigcup_{\mathcal{V}\in\mathcal{N}(\mathcal{V}(\mathbf{g}_{i, k}))}{hash}(\mathcal{V})$. Consequently, the optimal matched descriptor $\mathbf{g}^\star$ is formulated as: 
  \begin{equation}
  \begin{split}
 \mathbf{g}^\star = \underset{\mathbf{g}\in\mathcal{N}_{i, k}, v(\mathbf{g})\,\neq\, i}{\rm argmin}
	\|\mathbf{g}_{i, k} - \mathbf{g}\|_2,\ \text{s.t.}\|\mathbf{g}_{i, k} - \mathbf{g}\|_2 < \nu_c
  \end{split}
	\label{eq:neighborhood_nn}
\end{equation} 
where $v(\mathbf{g})$ returns the view index of the descriptor. The view constraint explicitly prevents intra-frame matching, while the distance threshold $\nu_c$ filters out putative correspondences with large residuals. 
Since the total number of candidates within $\mathcal{N}_{i, k}$ remains limited, it effectively reduces the overall matching complexity from $\mathcal{O}(S^2 N^2)$ to $\mathcal{O}(SN)$. In practice, we implement this neighborhood querying and matching process via parallelized custom CUDA kernels, further improving the cross-view matching efficiency. 

Consequently, building upon the efficient voxel hash-based descriptor matching above, we aggregate all identified cross-view correspondences into a consolidated set:
\begin{equation}
	\mathcal{C} = \Big\{\big(
	\mathbf{x}_k^{s},\;\mathbf{x}_k^{t},\;
	i_k,\;j_k,\;
	c_k^{s},\;c_k^{t}
	\big)\Big\}_{k=1}^{|\mathcal{C}|}
	\label{eq:corr_tuple}
\end{equation}
  where $k$ indexes the matched pairs; $i_k,\,j_k$ denote the respective view indices; $\mathbf{x}_k^{s} \in \mathbf{S}_{i_k}$ and
$\mathbf{x}_k^{t} \in \mathbf{S}_{j_k}$ represent the metric 3D coordinates retrieved from the 3D scans, and $c_k^{s},\,c_k^{t}$ are their corresponding per-point confidence scores predicted by the visual foundation models.

\subsubsection{Geometry-aware Robust Alignment Loss}
With the putative correspondence set $\mathcal{C}$ established in Sec.~\ref{corr},
we perform motion-only bundle adjustment to refine the initial camera poses
$(\hat{\mathbf{T}}_2^{(0)}, \hat{\mathbf{T}}_3^{(0)}, \ldots, \hat{\mathbf{T}}_S^{(0)})$ by jointly
minimizing geometry-aware alignment losses. 
It should be noted that to fix the gauge freedom of the pose-only optimization, we take the first frame
as the reference frame and keep its pose $\hat{\mathbf{T}}_1^{(0)}$ fixed, thereby anchoring the global
coordinate system.

Formally, we optimize the pose increments
$\boldsymbol{\xi} = (\boldsymbol{\xi}_2,\, \boldsymbol{\xi}_3,\, \ldots \, ,\boldsymbol{\xi}_S)$,
where each $\boldsymbol{\xi}_i=[\boldsymbol{\omega}_i^\trans, \boldsymbol{v}_i^\trans]^\trans \in \mathbb{R}^{6\times 1}$ denotes a local pose increment (i.e., a twist) in the Lie algebra $\mathfrak{se}(3)$, with $\boldsymbol{\omega}_i$ and $\boldsymbol{v}_i$ corresponding to the rotational and translational components, respectively. The refined $SE(3)$ pose
of view $i$ is parameterized as
\begin{equation}
\begin{split}
\hat{\mathbf{T}}_1 &= \hat{\mathbf{T}}_1^{(0)}, \\
\hat{\mathbf{T}}_i&=\boldsymbol{\xi}_i \oplus \hat{\mathbf{T}}_i^{(0)}=\operatorname{Exp}(\boldsymbol{\xi}_i)\,\hat{\mathbf{T}}_i^{(0)}, \quad i \in \{2, \dots, S\}
\end{split}
\end{equation} 
where $\hat{\mathbf{T}}_i^{(0)}$ is the initialization from Sec.~\ref{init}, and
$\oplus$ denotes the left-plus operator on the ${SE}(3)$ manifold;
$\operatorname{Exp}: \mathfrak{se}(3) \rightarrow {SE}(3)$ is the
exponential map that converts a Lie-algebra increment into a  rigid
transformation. 

For each correspondence $k \in \mathcal{C}$ between views $i_k$ and $j_k$, we
consider two complementary alignment cues, namely correspondence alignment and geometric reprojection alignment:

\noindent\textbf{Correspondence Alignment Residual.} 
The first residual measures the point-to-point alignment residual between the two matched 3D
points after transforming them into the world frame, which can be formulated as:
\begin{equation}
	\begin{split}
		\mathbf{r}_k^{{c}}(\boldsymbol{\xi})
		=
		\hat{\mathbf{T}}_{i_k}\mathbf{x}_k^s
		-
		\hat{\mathbf{T}}_{j_k}\mathbf{x}_k^t \in\mathbb{R}^{3\times 1},
	\end{split}
\end{equation}
where $\mathbf{x}_k^s$ and $\mathbf{x}_k^t$ denote the source and target 3D
points associated with the $k$-th correspondence; $i_k$ and $j_k$ denote the corresponding view indices, respectively.

\noindent\textbf{Geometric Reprojection Residual.}
We further employ the geometric reprojection alignment as the additional constraint cues to guide pose optimization. 
Its core idea is based on the fact that, under the correct poses, a source 3D point transformed into the target frame should coincide with the target-side 3D point reconstructed by sampling the target depth map at its reprojected image location. 
Formally, we project the source point
$\mathbf{x}_k^s$ into the target view $j_k$, sample the target depth map
$\mathbf{D}_{j_k}$ at the reprojected location, back-project the sampled depth
to 3D, and measure the geometric residual between the transformed source
point and the reconstructed target point: $\mathbf{r}_k^{{g}}(\boldsymbol{\xi})=	$
\begin{equation}
	\begin{split}
		{\hat{\mathbf{T}}_{j_k}}^{-1}\hat{\mathbf{T}}_{i_k}\mathbf{x}_k^s
		-
		\pi^{-1}
		\!\left(
		\mathbf{D}_{j_k}
		\!\left(
		\pi\!\left(
		\hat{\mathbf{T}}_{j_k}^{-1}\hat{\mathbf{T}}_{i_k}\mathbf{x}_k^s
		\right)
		\right)
		\right)\in\mathbb{R}^{3\times 1},
	\end{split}
\end{equation}
where $\pi(\cdot)$ and $\pi^{-1}(\cdot)$ denote the projection and back-projection
operators, respectively. 


\noindent\textbf{Robust Joint Optimization Objective.} Consequently, 
we fuse the formed correspondence alignment residual and  reprojection residual into a single joint residual:
\begin{equation}
	\begin{split}
     \mathbf{r}_k(\boldsymbol{\xi})=\begin{bmatrix} \sqrt{\lambda_c}\mathbf{r}_k^c(\boldsymbol{\xi})^\trans, \sqrt{\lambda_g}\mathbf{r}_k^g(\boldsymbol{\xi})^\trans\end{bmatrix}^\trans \in \mathbb{R}^{6\times 1}
	\end{split}
\end{equation}
where $\lambda_c$ and $\lambda_g$ denote the loss coefficients that control the relative contributions of each residual term. 
Building upon it, our robust joint  optimization objective over all correspondences can be formulated as:
\begin{equation}\label{joint}
	\begin{split}
		\mathcal{L}(\boldsymbol{\xi})=\sum_{k=1}^{|\mathcal{C}|}c_k\cdot\rho(\|\mathbf{r}_k(\boldsymbol{\xi})\|_2),
	\end{split}
\end{equation}
where $c_k=\sqrt{c_k^s \cdot c_k^t}$ denotes the {confidence} weight derived from the foundation model's per-point quality score for modulating each residual term. The higher confidence assigns larger optimization weights.  $\rho(x)=\frac{x^2/2}{1 + (x/\delta)^2}$ represents the Geman-McClure kernel, a robust function for suppressing the negative effectiveness of outlier correspondences with high residuals, where $\delta>0$ controls the kernel width. 

\subsubsection{IRLS-based Motion-only Bundle Adjustment}
Building upon the joint loss defined in Eq.~\ref{joint},  
our optimization objective is to find the optimal pose increments $\boldsymbol{\xi}^* =(\boldsymbol{\xi}_2^*, \boldsymbol{\xi}_3^*,... , \boldsymbol{\xi}_S^*)$ that can minimize the residual errors over all correspondences:
\begin{equation}\label{irls}
	\begin{split}
		\boldsymbol{\xi}^* =\arg\min_{\boldsymbol{\xi}}\mathcal{L}(\boldsymbol{\xi}) \triangleq \sum_{k=1}^{|\mathcal{C}|}c_k\cdot\rho(\|\mathbf{r}_k(\boldsymbol{\xi})\|_2). 
	\end{split}
\end{equation}
We follow the well-known Iteratively Reweighted Least Squares (IRLS)~\cite{holland1977robust} to solve this objective. At each iteration $t$, the robust kernel-based optimization objective is reformulated as the least-square form as follows:
\begin{equation}
	\begin{split}\label{loss}
		\boldsymbol{\xi}^{(t+1)} &=\arg\min_{\boldsymbol{\xi}}\sum_{k=1}^{|\mathcal{C}|} c_k\cdot w_k^{(t)}\cdot{\|\mathbf{r}_k(\boldsymbol{\xi})\|_2^2}	\end{split}
\end{equation}
where the robust weight $w_k^{(t)}$ is formulated as:
\begin{equation}
	\begin{split}
		w_k^{(t)} = \frac{\rho'(\|\mathbf{r}_k(\boldsymbol{\xi}^{(t)})\|_2)}{\|\mathbf{r}_k(\boldsymbol{\xi}^{(t)})\|_2} = \left(\frac{\delta^2}{\delta^2 + \|\mathbf{r}_k(\boldsymbol{\xi}^{(t)})\|_2^2}\right)^2.
	\end{split}
\end{equation} 
To solve the non-linear least-squares objective formulated in Eq.~\ref{loss}, we employ the Gauss-Newton algorithm. 
In detail, at each iteration $t$, we linearize the joint residual vector $\mathbf{r}_k(\boldsymbol{\xi})$ around the current pose estimate $\boldsymbol{\xi}^{(t)}$ via a first-order Taylor expansion: $\mathbf{r}_k(\boldsymbol{\xi}) \approx  \mathbf{r}_k(\boldsymbol{\xi}^{(t)}) + \mathbf{J}_k^{(t)} \Delta\boldsymbol{\xi}$, where $\Delta\boldsymbol{\xi}$ denotes the local left perturbation at the current iteration, and $\mathbf{J}_k^{(t)} = \left. \frac{\partial \mathbf{r}_k(\boldsymbol{\xi})}{\partial\boldsymbol{\xi}} \right|_{\boldsymbol{\xi}=\boldsymbol{\xi}^{(t)}}$ is the Jacobian matrix evaluated at the current state:
\begin{equation}\small
\begin{split}
\frac{\partial \mathbf{r}_k(\boldsymbol{\xi})}{\partial\boldsymbol{\xi}} = \begin{bmatrix}{
\sqrt{\lambda_c}\, \frac{\partial \mathbf{r}^c_k(\boldsymbol{\xi})}{\partial\boldsymbol{\xi}}} \\ \sqrt{\lambda_g}\, \frac{\partial \mathbf{r}^g_k(\boldsymbol{\xi})}{\partial\boldsymbol{\xi}}
\end{bmatrix}
=\begin{bmatrix}
\sqrt{\lambda_c}\, \mathbf{J}_k^c \\ 
\sqrt{\lambda_g}\, \mathbf{J}_k^g
\end{bmatrix}\in\mathbb{R}^{6\times 6\, (S-1)}.
\end{split}
\end{equation}
Here, $\mathbf{J}_k^{{c}}$ and $\mathbf{J}_k^{{g}}$ denote the Jacobian matrices of the correspondence residual and
geometric reprojection residual, respectively. Their
closed-form expressions are derived by standard first-order differentiation under
the left-plus perturbation on ${SE}(3)$ and are provided in Appendix~A. In particular, $\mathbf{J}_k^{{g}}$ is obtained by chaining the derivatives of rigid
transformation, perspective projection, and bilinear sampling. 
The IRLS optimization objective can then be rewritten as:
\begin{equation}\label{irls}
\begin{split}
\Delta\boldsymbol{\xi}^\star & = \arg\min_{\Delta\boldsymbol{\xi}} \sum_{k=1}^{|\mathcal{C}|}\beta_k^{(t)}\cdot\|\mathbf{r}_k(\boldsymbol{\xi}^{(t)}) + \mathbf{J}_k^{(t)} \Delta\boldsymbol{\xi}\|_2^2,
\end{split}
\end{equation}
where $\beta_k^{(t)} = c_k\cdot w_k^{(t)}$. 
By setting the partial derivatives to zero, the normal equation can be achieved:
\begin{equation}\label{normal}\small
	\begin{split}
\underbrace{\left(\sum_{k=1}^{|\mathcal{C}|}\beta_k^{(t)}{\mathbf{J}_k^{(t)}}^\trans {\mathbf{J}_k^{(t)}}\right)}_{\mathbf{H}}\Delta\boldsymbol{\xi}^\star = -\underbrace{\sum_{k=1}^{|\mathcal{C}|}\beta_k^{(t)}{\mathbf{J}_k^{(t)}}^\trans \mathbf{r}_k(\boldsymbol{\xi}^{(t)})}_{\mathbf{g}}.
	\end{split}
\end{equation}
Here, ${\mathbf{J}_k^\trans {\mathbf{J}}_k}$ and ${\mathbf{J}_k^\trans {\mathbf{r}}_k}$ can be decoupled as:
\begin{equation}
\begin{split}
{\mathbf{J}_k^\trans {\mathbf{J}}_k} &= \lambda_c\, {\mathbf{J}_k^c}^\trans {\mathbf{J}_k^c} + \lambda_g\, {\mathbf{J}_k^g}^\trans {\mathbf{J}_k^g}, \\
{\mathbf{J}_k^\trans {\mathbf{r}}_k} &= \lambda_c\, {\mathbf{J}_k^c}^\trans {\mathbf{r}_k^c} + \lambda_g\, {\mathbf{J}_k^g}^\trans {\mathbf{r}_k^g}.
\end{split}
\end{equation}

To efficiently solve the normal equation in Eq.~\ref{normal}, we employ a damped 
conjugate gradient as the solver. 
Specifically, at each
Gauss-Newton iteration, we solve the damped normal equation as follows:
\begin{equation}
	\left(\mathbf{H} + \lambda \mathbf{I}\right)
	\Delta\boldsymbol{\xi}^{\star}
	=
	-\mathbf{g},
\end{equation}
where $\lambda > 0$ is a small damping coefficient for improving numerical
stability. Finally, we leverage the conjugate gradient to iteratively update the search direction until the relative residual falls below a preset threshold.

\begin{figure*}[p]
	\centering
	\includegraphics[width=\textwidth]{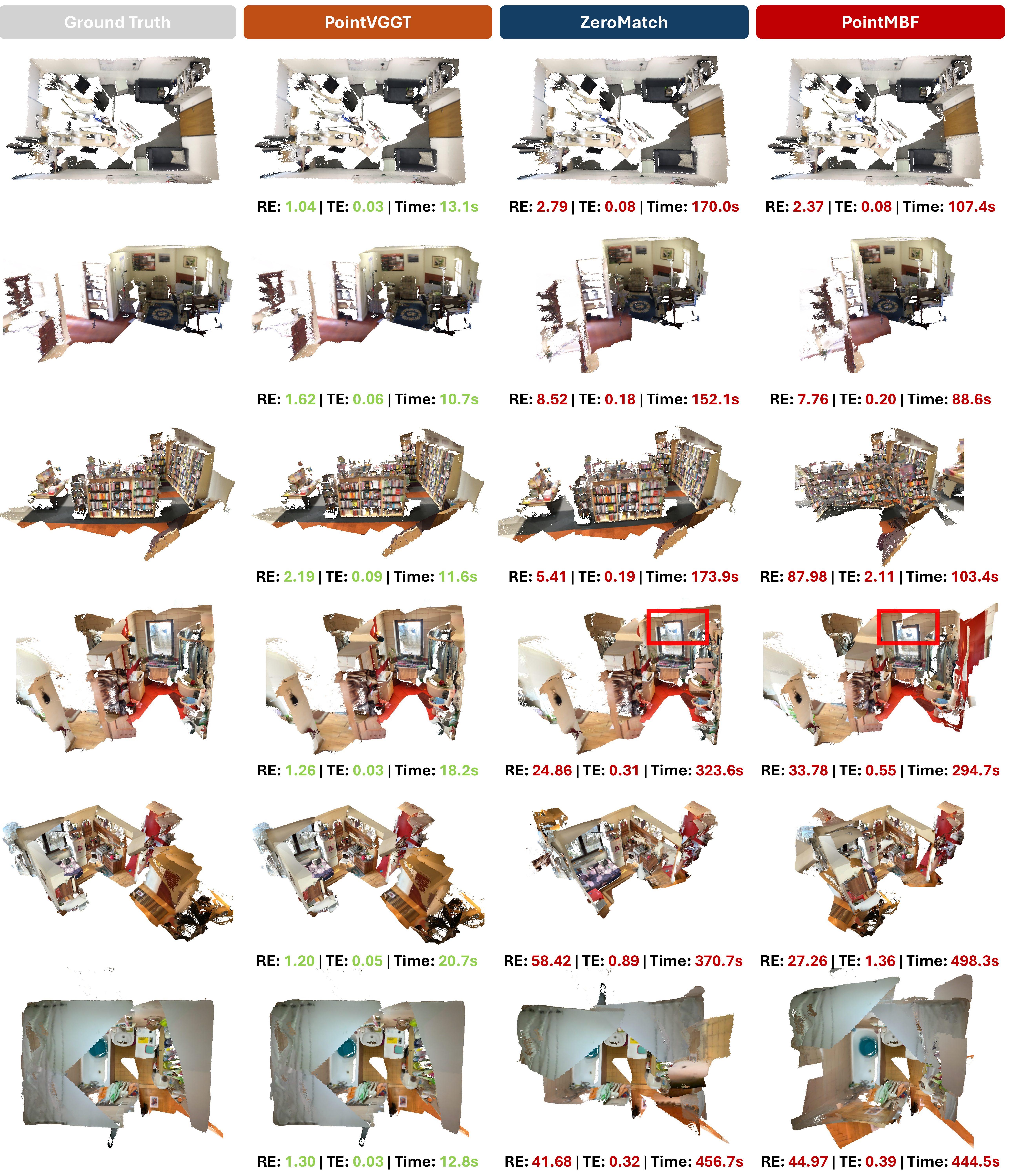}
	\caption{\textbf{Qualitative comparisons on ScanNet (top three rows) and ARKitScenes (bottom three rows)}.
		PointVGGT achieves more accurate alignment than two state-of-the-art RGB-D pairwise registration methods, ZeroMatch~\cite{Jiang_2025_CVPR} and PointMBF~\cite{yuan2023pointmbf} with PGO~\cite{Zhou2018} (i.e., pose graph optimization of Open3D).
		It should be highlighted that our proposed \emph{foundation-then-refinement} paradigm significantly reduces inference time from minutes to seconds compared with the conventional \emph{pairwise-then-global} pipeline.}
	\label{vis10}
\end{figure*}

\section{Experiments}
\subsection{Experimental Settings} \label{setting}
\noindent\textbf{Implementation Details.} 
PointVGGT is instantiated with three recently released VGGT-style foundation models: \textbf{VGGT}~\cite{wang2025vggt}, \textbf{Pi3}~\cite{wang2025pi}, and \textbf{DepthAnything3 (DA3)}~\cite{lin2025depth}. In the refinement stage, the voxel cell size $\nu$, the correspondence threshold $\nu_c$ and the robust kernel width $\delta$ are set to (0.015\,m, 0.045\,m, 0.05) for indoor scenes, and (0.05\,m, 0.15\,m, 0.1) for outdoor scenes. 
The loss coefficients $\lambda_c$ and $\lambda_g$ are set to 1.0 and 0.5, respectively. 
For the refinement solver, we set the learning rate to 0.005 and run the optimization for up to 40 iterations, with early stopping triggered when the relative change falls below $1\times 10^{-5}$. 
All experiments are conducted  on a single NVIDIA L20 GPU.


\noindent\textbf{Evaluation Metrics.}
To quantify registration accuracy, we measure deviations between estimated relative poses $\{\hat{\mathbf{R}}_{ij}, \hat{\mathbf{t}}_{ij}\}$ and their ground-truth counterparts $\{\mathbf{R}_{ij}, \mathbf{t}_{ij}\}$. Consistent with established evaluation protocols~\cite{wang2023robust,yew2021learning,gojcic2020learning}, the geometric discrepancies are decoupled into angular rotation error ($\operatorname{RE}_{ij}$) and spatial translation error ($\operatorname{TE}_{ij}$) for every evaluated scan pair $(i, j)$:
\begin{equation}
\begin{split}
&\operatorname{RE}_{ij} = \operatorname{arccos} \frac{\operatorname{Tr}(\hat{\mathbf{R}}_{ij}^\top \mathbf{R}_{ij}) - 1}{2}, \\
&\operatorname{TE}_{ij} = \|\hat{\mathbf{t}}_{ij} - \mathbf{t}_{ij}\|_2.
\end{split}
\end{equation}
We adopt the Empirical Cumulative Distribution Function (ECDF)~\cite{gojcic2020learning} to characterize the proportion of successfully registered pairs within varying error margins $\gamma$. For a universal set of evaluated pairs $\mathcal{P}$, the ECDF at tolerance $\gamma$ is formulated as:
\begin{equation}
    \operatorname{ECDF}(\gamma) = \frac{\big| \{ (i,j) \in \mathcal{P} \mid \mathcal{E}_{ij} \leq \gamma \} \big|}{|\mathcal{P}|},
\end{equation}
where $\mathcal{E}_{ij}$ denotes either $\operatorname{RE}_{ij}$ or $\operatorname{TE}_{ij}$, and $|\mathcal{P}|$ is the total number of evaluated pairs. For rotation, recall is reported at different rotation and translation thresholds. 
 Mean and median errors are additionally reported to capture the overall distribution of registration quality beyond fixed thresholds. This joint set of metrics enables a comprehensive assessment of both precision (strict thresholds) and robustness (relaxed thresholds) under varying scene difficulty levels.

\begin{table*}[h]
	\centering
	\caption{Performance comparisons on indoor \textbf{ScanNet}~\cite{dai2017scannet}, \textbf{ScanNet++}~\cite{yeshwanth2023scannet++} and \textbf{ArkitScenes}~\cite{baruch2021arkitscenes} benchmark datasets.}
	\resizebox{1\textwidth}{!}{
		\begin{tabular}{cclcccccccccccccc}
			\toprule[1.8pt]
			\multirow{3}{*}{\textbf{Dataset}} &\multirow{3}{*}{\textbf{Global}} & \multirow{3}{*}{\textbf{Method}}& \multicolumn{6}{c}{\textbf{Rotation Error}} & \multicolumn{6}{c}{\textbf{Translation Error} (\textit{m})} \\
			\cmidrule(lr){4-9} \cmidrule(lr){10-15} 
			& & & 3$^\circ$ & 5$^\circ$ & 10$^\circ$ & 30$^\circ$ & 45$^\circ$ & Mean/Med & 0.05 & 0.1 & 0.25 & 0.5 & 0.75 & Mean/Med  \\
            \midrule[1.6pt]
            \multirow{18}{*}{\textbf{ScanNet}} & \multirow{5}{*}{\rotatebox[origin=c]{90}{\textbf{GTSAM}}} & LLT & 12.4 & 16.3 & 24.6 & 44.2 & 52.1 & 53.9/38.7 &  7.3 & 13.8 & 26.2 & 37.6 & 48.1 & 1.19/0.80 \\
            & & UR\&R & 31.4 & 42.3 & 52.4 & 61.1 & 63.7 & 43.6/8.3 & 21.0 & 29.1 & 46.5 & 56.7 & 62.0 & 1.02/0.30 \\
            & & PointMBF & 36.4 & 44.5 & 49.6 & 54.3 & 61.5 & 46.8/12.4 & 24.8 & 37.5 & 48.2 & 51.4 & 55.8 & 1.05/0.42 \\
            & & ColorPCR & 48.7 & 66.7 & 76.3 & 78.3 & 78.6 & 35.4/3.1 & 27.6 & 49.8 & 74.2 & 77.3 & 77.4 & 0.68/0.10 \\
            & & ZeroMatch & 40.7 & 48.8 & 56.5 & 57.8 & 60.3 & 47.8/5.4 & 28.2 & 43.7 & 48.5 & 52.3 & 55.8 & 0.94/0.38 \\
            \cmidrule(lr){2-15}
			&\multirow{5}{*}{\rotatebox[origin=c]{90}{\textbf{SGHR}}} & LLT & 13.3 & 19.2 & 30.8 & 48.8 & 55.5 & 50.7/31.9 &  7.6 & 15.0 & 29.1 & 42.6 & 51.8 & 1.22/0.70 \\
            && UR\&R & 33.8 & 47.2 & 54.4 & 58.8 & 61.1 & 48.2/5.7 &23.6 & 36.8 & 52.7 & 56.8 & 60.5 & 1.11/0.17 \\
            && PointMBF & 37.1 & 48.0 & 54.1 & 61.9 & 64.2 & 43.3/5.5 & 26.2 & 40.1 & 51.2 & 54.8 & 57.0 & 1.06/0.21 \\
            && ColorPCR & 50.9 & 67.9 & 78.2 & 81.1 & 83.7 & 28.7/2.9 & 26.5 & 47.2 & 71.7 & 78.3 & 78.8 & 0.74/0.11 \\
            && ZeroMatch & 51.7 & 64.0 & 70.6 & 73.2 & 73.8 & 34.2/2.9 & 32.8 & 54.9 & 66.7 & 71.6 & 74.9 & 0.64/0.08 \\
            \cmidrule(lr){2-15}
			&\multirow{5}{*}{\rotatebox[origin=c]{90}{\textbf{PGO}}} & LLT &  16.2 & 25.8 & 43.5 & 73.8 & 82.7 & 29.8/11.9 & 11.0 & 21.0 & 38.6 & 57.6 & 67.0 & 0.74/0.40 \\
            && UR\&R &  45.6 & 62.7 & 76.4 & 89.6 & 91.9 & 15.7/3.4 & 27.8 & 49.8 & 68.5 & 82.4 & 85.8 & 0.41/0.10 \\
            && PointMBF &  56.2 & 75.5 & 85.1 & 92.4 & 92.4 & 15.7/2.6 & 31.2 & 59.3 & 79.8 & 86.9 & 88.3 & 0.38/0.08  \\
            && ColorPCR & 51.2 & 70.6 & 91.3 & 95.4 & 95.8 & 11.2/2.9 & 27.3 & 49.5 & 81.5 & 86.5 & 87.1 & 0.48/0.10 \\
            && ZeroMatch & 57.4 & 75.2 & 88.5 & 97.4 & 97.4 & 6.2/2.5 & 39.1 & 63.7 & 82.0 & 93.0 & 95.4 & 0.15/0.07 \\
            \cmidrule(lr){2-15}
            && \textcolor[RGB]{166,35,35}{\textbf{PointVGGT}} (\textbf{VGGT}) &  24.5 & 48.0 & 87.1 & \textbf{100.0} & \textbf{100.0} & 5.8/5.2 & 14.5 & 37.0 & 83.2 & 97.7 & 99.8 & 0.16/0.13  \\
            && \textcolor[RGB]{166,35,35}{\textbf{PointVGGT}} (\textbf{DA3}) & 81.0 & 94.9 & \textbf{100.0} & \textbf{100.0} & \textbf{100.0} & 2.0/1.6 & 55.5 & 80.1 & 96.0 & 98.6 & 99.9 & 0.07/0.04 \\
			&& \textcolor[RGB]{166,35,35}{\textbf{PointVGGT}} (\textbf{PI3}) & \textbf{92.2} & \textbf{99.3} & \textbf{100.0} & \textbf{100.0} & \textbf{100.0} & \textbf{1.4}/\textbf{1.1} & \textbf{73.4} & \textbf{89.0} & \textbf{98.0} & \textbf{100.0} & \textbf{100.0} & \textbf{0.05}/\textbf{0.03}\\
            \midrule[1.6pt]
            \multirow{18}{*}{\textbf{ScanNet++}} &\multirow{5}{*}{\rotatebox[origin=c]{90}{\textbf{GTSAM}}} & LLT & 43.0 & 55.3 & 75.6 & 94.7 & 99.5 & 7.7/3.8 & 26.3 & 43.9 & 68.9 & 84.9 & 93.2 & 0.24/0.13 \\
            && UR\&R & 75.8 & 82.5 & 89.3 & 99.7 & 99.7 & 3.4/1.5 & 59.9 & 79.9 & 86.9 & 95.1 & 97.9 & 0.11/0.04 \\
            && PointMBF & 81.1 & 88.4 & 93.6 & 99.7 & 99.7 & 3.0/1.2 & 69.0 & 83.3 & 90.0 & 96.3 & 98.3 & 0.09/0.03 \\
            && ColorPCR & 71.3 & 86.4 & 99.0 & \textbf{100.0} & \textbf{100.0} & 2.5/1.6 & 39.6 & 66.4 & 86.7 & 96.9 & 98.9 & 0.122/0.065 \\
            && ZeroMatch & 91.2 & 96.1 & 98.0 & \textbf{100.0} & \textbf{100.0} & 1.7/1.2 & 68.4 & 82.1 & 87.3 & 94.2 & 98.9 & 0.10/0.03 \\
            \cmidrule(lr){2-15}
			&\multirow{5}{*}{\rotatebox[origin=c]{90}{\textbf{SGHR}}} & LLT & 42.3 & 56.6 & 74.7 & 95.2 & 99.4 & 8.3/3.9 & 26.7 & 44.3 & 64.8 & 80.6 & 90.8 & 0.26/0.13 \\
            && UR\&R & 76.1 & 82.7 & 89.3 & \textbf{100.0} & \textbf{100.0} & 3.3/1.5 & 58.7 & 78.2 & 87.3 & 94.6 & 98.1 & 0.11/0.04 \\
            && PointMBF & 80.0 & 83.7 & 87.5 & 94.4 & 95.0 & 10.6/1.3 & 67.4 & 78.8 & 84.8 & 91.0 & 93.2 & 0.20/0.03 \\
            && ColorPCR & 72.1 & 85.5 & 96.9 & 98.0 & 98.0 & 4.9/1.6 & 40.0 & 66.3 & 85.5 & 93.3 & 96.6 & 0.19/0.06 \\
            && ZeroMatch & 90.9 & 96.4 & 98.6 & \textbf{100.0} & \textbf{100.0} & 1.7/1.2 & 66.0 & 82.6 & 88.6 & 95.6 & 99.1 & 0.09/0.04 \\
            \cmidrule(lr){2-15}
			&\multirow{5}{*}{\rotatebox[origin=c]{90}{\textbf{PGO}}} & LLT & 35.3 & 51.6 & 73.6 & 96.5 & 99.3 & 7.9/4.8 & 23.8 & 44.0 & 66.6 & 82.2 & 90.4 & 0.29/0.12  \\
            && UR\&R & 69.5 & 75.2 & 83.8 & 98.2 & 98.6 & 6.2/1.6 & 55.2 & 70.6 & 81.4 & 91.1 & 97.1 & 0.16/0.04 \\
            && PointMBF &  80.0 & 88.2 & 94.7 & 99.6 & \textbf{100.0} & 2.7/1.3 & 68.4 & 83.3 & 92.8 & 98.1 & 99.5 & 0.07/0.03  \\
            && ColorPCR & 66.5 & 78.5 & 91.0 & 95.7 & 95.7 & 9.3/1.8 & 40.2 & 66.5 & 86.3 & 91.9 & 94.2 & 0.51/0.06 \\
            && ZeroMatch & 89.4 & 97.0 & 98.7 & \textbf{100.0} & \textbf{100.0} & 1.7/1.3 & 70.5 & 84.7 & 91.4 & 98.6 & 99.5 & 0.07/0.03 \\
            \cmidrule(lr){2-15}
            && \textcolor[RGB]{166,35,35}{\textbf{PointVGGT}} (\textbf{VGGT}) & 90.5 & 97.6 & \textbf{100.0} & \textbf{100.0} & \textbf{100.0} & 1.4/1.1 & 77.3 & 94.3 & \textbf{100.0} & \textbf{100.0} & \textbf{100.0} & 0.04/0.03 \\
            && \textcolor[RGB]{166,35,35}{\textbf{PointVGGT}} (\textbf{DA3}) &  \textbf{100.0} & \textbf{100.0} & \textbf{100.0} & \textbf{100.0} & \textbf{100.0} & \textbf{0.7}/\textbf{0.6} & \textbf{95.4} & \textbf{99.9} & \textbf{100.0} & \textbf{100.0} & \textbf{100.0} & \textbf{0.02}/\textbf{0.02} \\
			&& \textcolor[RGB]{166,35,35}{\textbf{PointVGGT}} (\textbf{PI3}) & 94.8 & 96.2 & 99.4 & \textbf{100.0} & \textbf{100.0} & 1.1/0.7 & 90.8 & \textbf{99.9} & \textbf{100.0} & \textbf{100.0} & \textbf{100.0} & 0.03/\textbf{0.02} \\
            \midrule[1.6pt]
            \multirow{18}{*}{\textbf{ArkitScenes}}&\multirow{5}{*}{\rotatebox[origin=c]{90}{\textbf{GTSAM}}} & LLT & 2.7 &  4.4 &  8.1 & 20.4 & 31.8 & 82.7/78.7 &  2.4 &  4.7 &  9.8 & 21.4 & 34.0 & 1.22/1.01 \\
            && UR\&R & 7.9 & 11.1 & 17.3 & 29.0 & 35.6 & 84.0/84.9 &  6.6 & 10.5 & 17.4 & 24.3 & 34.0 & 1.40/1.11 \\
            && PointMBF & 8.7 & 10.9 & 16.4 & 26.7 & 31.7 & 81.4/81.9 &  7.1 & 10.2 & 15.9 & 25.8 & 34.5 & 1.39/1.15 \\
            && ColorPCR & 19.0 & 28.4 & 37.9 & 44.8 & 46.0 & 68.5/57.5 & 15.0 & 25.6 & 37.0 & 45.6 & 54.1 & 1.01/0.63 \\
            && ZeroMatch &  19.3 & 26.8 & 29.5 & 33.4 & 37.0 & 77.2/86.8 & 11.9 & 15.4 & 19.2 & 28.6 & 42.7 & 1.10/0.90 \\
            \cmidrule(lr){2-15}
			&\multirow{5}{*}{\rotatebox[origin=c]{90}{\textbf{SGHR}}} & LLT & 2.4 &  4.1 &  7.2 & 18.7 & 26.0 & 91.2/94.0 &  2.1 &  4.4 &  8.8 & 16.0 & 26.1 & 1.56/1.26 \\
            && UR\&R & 6.9 &  9.2 & 13.6 & 23.7 & 27.5 & 91.6/104.2 &  6.2 &  9.8 & 16.5 & 24.2 & 32.4 & 1.59/1.28 \\
            && PointMBF & 6.6 &  8.9 & 12.0 & 14.4 & 18.6 & 104.2/115.6 &  4.9 &  8.4 & 13.5 & 18.3 & 25.2 & 1.57/1.37 \\
            && ColorPCR & 14.7 & 20.5 & 26.1 & 31.5 & 34.7 & 86.8/92.2 & 12.1 & 19.4 & 27.7 & 35.2 & 42.7 & 1.51/1.05 \\
            && ZeroMatch & 28.1 & 39.8 & 47.7 & 48.7 & 49.2 & 62.4/78.9 & 17.5 & 29.0 & 36.7 & 44.9 & 56.5 & 0.97/0.58 \\
            \cmidrule(lr){2-15}
			&\multirow{5}{*}{\rotatebox[origin=c]{90}{\textbf{PGO}}} & LLT & 3.2 &  6.3 & 12.5 & 34.9 & 44.9 & 71.3/59.4 &  2.7 &  6.5 & 15.0 & 27.5 & 38.0 & 1.46/1.14 \\
            && UR\&R  & 10.0 & 14.4 & 22.4 & 38.0 & 41.5 & 83.9/75.5 &7.3 & 13.1 & 23.8 & 34.8 & 40.1 & 1.83/1.24 \\
            && PointMBF & 14.8 & 20.9 & 31.6 & 51.5 & 58.8 & 60.9/27.7 & 10.4 & 19.0 & 37.2 & 48.2 & 54.5 & 1.21/0.57   \\
            && ColorPCR & 18.0 & 29.4 & 37.7 & 48.3 & 49.4 & 73.3/64.2 &14.9 & 28.1 & 38.9 & 43.9 & 47.3 & 1.45/0.99 \\
            && ZeroMatch &  31.3 & 48.5 & 60.1 & 61.7 & 62.2 & 48.6/5.3 & 24.2 & 40.5 & 51.5 & 57.0 & 64.6 & 0.86/0.21 \\
            \cmidrule(lr){2-15}
            && \textcolor[RGB]{166,35,35}{\textbf{PointVGGT}} (\textbf{VGGT}) & 30.2 & 54.7 & 81.0 & 91.1 & 93.8 & 11.4/4.6 & 21.3 & 48.7 & 79.4 & 88.8 & 93.0 & 0.24/0.10 \\
            && \textcolor[RGB]{166,35,35}{\textbf{PointVGGT}} (\textbf{DA3}) & \textbf{75.2} & \textbf{77.6} & 80.4 & 85.2 & 87.8 & 15.1/\textbf{1.2} & \textbf{59.9} & \textbf{76.2} & 81.8 & 85.1 & 88.2 & 0.26/\textbf{0.04} \\
			&& \textcolor[RGB]{166,35,35}{\textbf{PointVGGT}} (\textbf{PI3}) & 72.1 & 77.2 & \textbf{83.9} & \textbf{96.1} & \textbf{99.2} & \textbf{5.4}/{1.5} & 52.4 & 73.3 & \textbf{84.4} & \textbf{93.9} & \textbf{97.8} & \textbf{0.13}/{0.05}
 \\
			\bottomrule[1.8pt]
	\end{tabular}}
	\label{allindoor}
\end{table*}

\begin{table*}[h]
	\centering
	\caption{Performance comparisons on object-centric \textbf{WildRGB-D}~\cite{xia2024rgbd} benchmark dataset.}
	\resizebox{1\textwidth}{!}{
		\begin{tabular}{clcccccccccccccc}
			\toprule[1.8pt]
			\multirow{3}{*}{\textbf{Pose Graph}} & \multirow{3}{*}{\textbf{Method}}& \multicolumn{6}{c}{\textbf{Rotation Error}} & \multicolumn{6}{c}{\textbf{Translation Error} (\textit{m})} \\
			\cmidrule(lr){3-8} \cmidrule(lr){9-14} 
			&  & 3$^\circ$ & 5$^\circ$ & 10$^\circ$ & 30$^\circ$ & 45$^\circ$ & Mean/Med & 0.05 & 0.1 & 0.25 & 0.5 & 0.75 & Mean/Med  \\
            \midrule[1.2pt]
            \multirow{5}{*}{\textbf{GTSAM}} & LLT & 11.3 & 14.3 & 17.4 & 26.2 & 32.8 & 83.3/85.9 & 13.3 & 18.5 & 28.3 & 48.3 & 66.5 & 0.56/0.53 \\
            & UR\&R & 18.7 & 22.1 & 24.2 & 28.4 & 33.8 & 83.2/91.5 & 17.9 & 22.7 & 30.0 & 44.3 & 67.6 & 0.55/0.56 \\
            & PointMBF &  6.9 & 10.7 & 15.5 & 23.8 & 29.4 & 84.5/86.3 & 10.1 & 14.8 & 24.8 & 46.5 & 67.9 & 0.57/0.52 \\
            & ColorPCR & 25.4 & 39.3 & 51.0 & 60.7 & 64.5 & 49.2/9.2 & 32.7 & 47.1 & 60.7 & 79.0 & 92.2 & 0.27/0.12 \\
            & ZeroMatch & 15.6 & 19.7 & 21.6 & 25.7 & 28.7 & 84.2/85.6 & 16.4 & 20.2 & 25.9 & 41.8 & 66.4 & 0.59/0.57 \\
             \cmidrule(lr){1-14}
			\multirow{5}{*}{\textbf{SGHR}} & LLT & 11.7 & 14.4 & 17.9 & 27.4 & 34.0 & 83.3/84.9 & 13.3 & 17.9 & 29.6 & 47.7 & 66.1 & 0.56/0.53 \\
            & UR\&R & 22.2 & 26.3 & 29.1 & 36.8 & 39.3 & 78.0/82.4 & 19.8 & 28.0 & 37.2 & 49.4 & 68.1 & 0.51/0.51 \\
            & PointMBF & 6.8 & 10.6 & 15.4 & 26.8 & 33.4 & 84.7/86.4 & 10.6 & 16.2 & 26.4 & 47.5 & 69.3 & 0.57/0.52 \\
            & ColorPCR & 27.5 & 38.1 & 53.4 & 59.6 & 61.7 & 64.7/8.7 & 28.8 & 46.2 & 59.3 & 70.3 & 87.4 & 0.36/0.12\\
            & ZeroMatch & 25.4 & 35.1 & 41.2 & 46.7 & 49.0 & 65.2/54.2 & 23.5 & 32.4 & 42.6 & 60.0 & 72.2 & 0.48/0.39 \\
            \cmidrule(lr){1-14}
			\multirow{5}{*}{\textbf{PGO}} & LLT & 26.5 & 40.5 & 57.5 & 87.0 & 92.4 & 18.0/7.1 & 35.9 & 48.7 & 76.4 & 88.7 & 92.2 & 0.20/0.10 \\
            & UR\&R  & 58.3 & 76.0 & 88.5 & 94.5 & 94.5 & 10.1/2.5 & 55.5 & 68.2 & 86.4 & 93.9 & 95.1 & 0.13/0.04 \\
            & PointMBF & 10.2 & 18.3 & 36.2 & 67.6 & 76.1 & 34.5/16.3 & 25.8 & 40.3 & 63.1 & 79.5 & 87.9 & 0.30/0.16 \\
            & ColorPCR & 20.1 & 29.6 & 43.5 & 54.0 & 59.9 & 64.8/22.6 & 22.4 & 35.7 & 48.2 & 69.6 & 85.6 & 0.35/0.27 \\
            & ZeroMatch & 39.7 & 57.8 & 73.1 & 85.6 & 89.7 & 22.0/4.1 & 37.4 & 54.4 & 75.9 & 87.5 & 90.6 & 0.21/0.09 \\
             \cmidrule(lr){1-14}
            & \textcolor[RGB]{166,35,35}{\textbf{PointVGGT}} (\textbf{VGGT}) & 31.4 & 59.8 & 92.9 & \textbf{100.0} & \textbf{100.0} & 4.7/4.3 & 71.9 & 95.7 & 99.1 & \textbf{100.0} & \textbf{100.0} & 0.04/0.03 \\
            & \textcolor[RGB]{166,35,35}{\textbf{PointVGGT}} (\textbf{DA3}) & 80.8 & 93.1 & 98.5 & \textbf{100.0} & \textbf{100.0} & 2.0/1.4 & 92.9 & 98.8 & \textbf{100.0} & \textbf{100.0} & \textbf{100.0} & 0.02/\textbf{0.01} \\
			& \textcolor[RGB]{166,35,35}{\textbf{PointVGGT}} (\textbf{PI3}) & \textbf{88.7} & \textbf{93.5} & \textbf{98.9} & \textbf{100.0} & \textbf{100.0} & \textbf{1.4}/\textbf{0.8} & \textbf{93.4} & \textbf{99.2} & \textbf{100.0} & \textbf{100.0} & \textbf{100.0} & \textbf{0.01}/\textbf{0.01} \\
			\bottomrule[1.8pt]
	\end{tabular}}
	\label{wild}
\end{table*}

\begin{table*}[h]
	\centering
	\caption{Performance comparisons  on outdoor \textbf{KITTI}~\cite{geiger2012we} and \textbf{Waymo}~\cite{sun2020scalability} benchmark datasets.}
	\resizebox{1\textwidth}{!}{
		\begin{tabular}{lcccccc|cccccc}
			\toprule[1.8pt]
			 \multirow{3}{*}{\textbf{Method}}& \multicolumn{3}{c}{\textbf{Rotation Error}} & \multicolumn{3}{c}{\textbf{Translation Error} (\textit{m})} & \multicolumn{3}{c}{\textbf{Rotation Error}} & \multicolumn{3}{c}{\textbf{Translation Error} (\textit{m})}  \\
			\cmidrule(lr){2-4} \cmidrule(lr){5-7} \cmidrule(lr){8-10}  \cmidrule(lr){11-13} 
			& 5$^\circ$ & 10$^\circ$ & Mean/Med  & 0.5 & 0.75 & Mean/Med  & 5$^\circ$ & 10$^\circ$ & Mean/Med  & 0.5 & 0.75 & Mean/Med \\
            \midrule[1.6pt]
            ZeroMatch + GTSAM & 94.6 & 95.1 & 1.2/0.2 & 69.8 & 72.7 & 2.18/0.14 & \textbf{100.0} & \textbf{100.0} & \textbf{0.1}/\textbf{0.1} & 69.3 & 69.9 & 3.14/0.06 \\
            ZeroMatch + SGHR &  95.7 & 98.0 & 0.7/0.2 &  52.4 & 57.7 & 2.22/0.37 &  \textbf{100.0} & \textbf{100.0} & \textbf{0.1}/\textbf{0.1} & 58.4 & 61.6 & 3.15/0.11 \\
            ZeroMatch + PGO &  \textbf{100.0} & \textbf{100.0} & 0.5/0.3 &  64.9 & 73.6 & 0.99/0.22 & \textbf{100.0} & \textbf{100.0} & 0.5/0.3 & 75.7 & 76.4 & 1.91/0.06 \\
            \cmidrule(lr){1-13}
            \textcolor[RGB]{166,35,35}{\textbf{PointVGGT}} (\textbf{VGGT}) &  90.0 & 98.5 & 1.7/0.6 & 58.8 & 73.1 & 0.53/0.34 & \textbf{100.0} & \textbf{100.0} & \textbf{0.1}/0.6 & \textbf{98.7} & \textbf{99.7} & \textbf{0.07}/\textbf{0.03} \\
            \textcolor[RGB]{166,35,35}{\textbf{PointVGGT}} (\textbf{DA3}) &  90.3 & 97.8  & 1.7/0.5 & \textbf{80.9} & \textbf{88.6} & \textbf{0.35}/\textbf{0.19} & \textbf{100.0} & \textbf{100.0} & \textbf{0.1}/\textbf{0.1} & 95.2 & 98.0 & 0.09/0.04 \\
			\textcolor[RGB]{166,35,35}{\textbf{PointVGGT}} (\textbf{PI3}) &  \textbf{100.0} & \textbf{100.0} & \textbf{0.3}/\textbf{0.2} &  75.6 & 85.0 & 0.41/0.24 & \textbf{100.0} & \textbf{100.0} & \textbf{0.1}/\textbf{0.1} & 84.6 & 89.6 & 0.25/0.08 \\
			\bottomrule[1.8pt]
	\end{tabular}}
	\label{scan}
\end{table*}

\subsection{Comparison with Existing Methods}
\subsubsection{Evaluation on Indoor Datasets}
\noindent\textbf{ScanNet.}
ScanNet~\cite{dai2017scannet} is a large-scale indoor RGB-D benchmark captured with structured-light depth sensors, covering a wide variety of room types and furniture configurations. Following the standard protocol~\cite{wang2023robust}, we evaluate on 14 test scenes, each containing 30 point clouds sampled at 20-frame intervals. We compare against five state-of-the-art pairwise RGB-D registration methods: LLT~\cite{wang2022improving}, UR\&R~\cite{el2021unsupervisedr}, PointMBF~\cite{yuan2023pointmbf}, ColorPCR~\cite{mu2024colorpcr}, and ZeroMatch~\cite{Jiang_2025_CVPR}, each combined with three global synchronization backends: GTSAM~\cite{gtsam}, SGHR~\cite{wang2023robust}, and PGO (pose graph optimization from Open3D)~\cite{Zhou2018}, yielding 15 competitive baselines in total. 

As shown in the top block of Table~\ref{allindoor}, PointVGGT with Pi3 and DA3 consistently and substantially outperforms the full set of \textit{pairwise-then-global} baselines. Among these baselines, ZeroMatch+PGO represents the strongest competitor, achieving rotation recall of 57.4\%/75.2\%/88.5\% at 3$^\circ$/5$^\circ$/10$^\circ$ and a mean rotation error of 6.2$^\circ$. PointVGGT (Pi3) exceeds these numbers by a wide margin, that is 92.2\%/99.3\%/100.0\% recall at the same thresholds, reducing mean rotation error by $4.1\times$ to 1.4$^\circ$ and mean translation error from 0.15\,m to 0.05\,m. Notably, even PointVGGT instantiated with the preliminary VGGT achieves 100\% recall at 30$^\circ$/45$^\circ$. 
We attribute this strong performance to the holistic visual geometry priors provided by the foundation models, which jointly reason over all views. This fundamentally avoids the error propagation and outlier sensitivity inherent to conventional \textit{pairwise-then-global} pipelines. Additional qualitative comparisons are presented in Fig.~\ref{vis10} and Appendix~B.

\noindent\textbf{ScanNet++.}
ScanNet++~\cite{yeshwanth2023scannet++} consists of unconstrained indoor RGB-D sequences captured by iPhone LiDAR sensors. In contrast to the clean structured-light depth in ScanNet, the raw LiDAR depth in ScanNet++ is affected by severe measurement noise, motion blur, and incomplete depth observations, all of which significantly undermine the reliability of multiview matching. We evaluate on 14 sequences from the \texttt{nvs\_test\_iphone} subset, where keyframes are sampled every 20 frames and up to the first 60 keyframes of each sequence are retained, following the same baseline comparison protocol as in ScanNet. 
As shown in the middle block of Table~\ref{allindoor}, PointVGGT (DA3) achieves impressive registration precision, yielding 100\% recall under all evaluated rotation and translation thresholds, with mean/median rotation errors of 0.7$^\circ$/0.6$^\circ$ and mean/median translation errors of 0.02\,m/0.02\,m. Compared with the strongest baseline, ZeroMatch+PGO, PointVGGT (DA3) improves the 3$^\circ$ rotation recall by 10.6 points and the 0.05\,m translation recall by 24.9 points. The gap becomes even larger under stricter thresholds, showing that PointVGGT remains robust even when noisy and incomplete depth severely degrades feature matching.

\noindent\textbf{ARKitScenes.}
ARKitScenes~\cite{baruch2021arkitscenes} is a large-scale indoor RGB-D dataset captured with iPhone LiDAR. We uniformly sample keyframes at a stride of 5 and retain up to the first 80 keyframes of each sequence. The bottom block of Table~\ref{allindoor} shows that the strongest baseline, ZeroMatch+PGO, achieves only 31.3\%/48.5\%/60.1\% rotation recall at 3$^\circ$/5$^\circ$/10$^\circ$, underscoring the difficulty of this benchmark for conventional two-stage pipelines. In contrast, PointVGGT (Pi3) reaches 72.1\%/77.2\%/83.9\% at the same thresholds, demonstrating robust global pose recovery under severely degraded depth conditions. PointVGGT (DA3) achieves the best performance at the strict 3$^\circ$/5$^\circ$ thresholds, with 75.2\%/77.6\% recall, while Pi3 performs better at the relaxed 30$^\circ$/45$^\circ$ thresholds (96.1\%/99.2\% vs.\ 85.2\%/87.8\% for DA3), suggesting complementary strengths across different foundation backbones. Together with the results on ScanNet and ScanNet++, these findings confirm the generality of the proposed foundation-then-refinement paradigm across diverse indoor sensors, ranging from clean structured-light cameras to raw consumer LiDAR.

\subsubsection{Evaluation on Object-centric Benchmark}
\begin{figure*}[h]
	\centering
	\includegraphics[width=\textwidth]{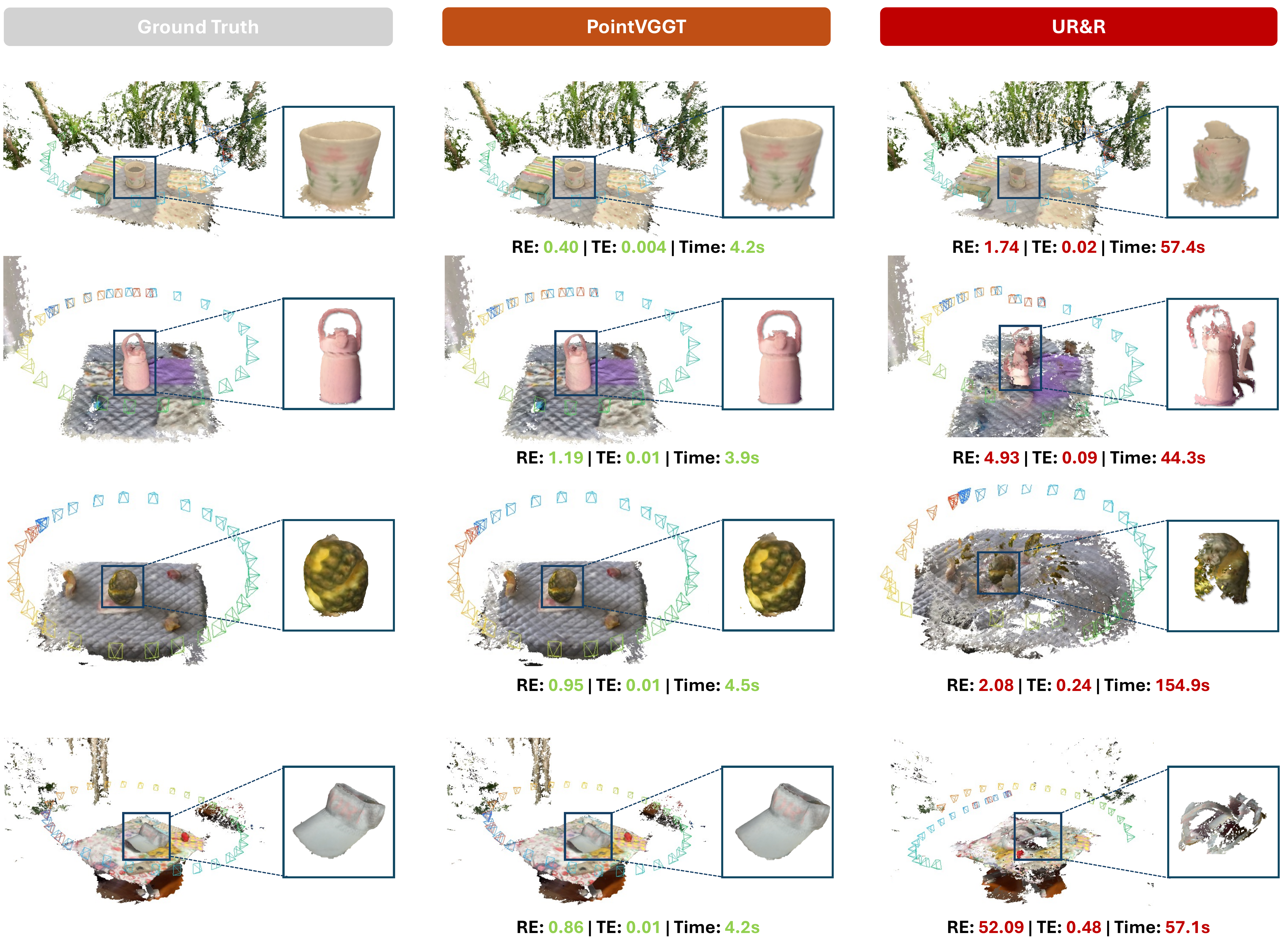}
	\caption{Qualitative comparisons on object-centric scenes from \textbf{WildRGB-D}~\cite{xia2024rgbd}. Our PointVGGT demonstrates impressive object reconstruction quality, producing more coherent and accurate surfaces and shapes than UR\&R+PGO.}
	\label{wildvis}
\end{figure*}

\noindent\textbf{WildRGB-D.}
WildRGB-D~\cite{xia2024rgbd} is a large-scale object-centric RGB-D dataset collected from 360$^\circ$ iPhone videos of diverse real-world objects in cluttered environments. Compared with scene-level registration, it poses distinct challenges, including small object scales, severe background clutter, diverse geometric structures, and full 360$^\circ$ viewpoint coverage under handheld capture. For evaluation, we randomly select 17 sequences, each corresponding to a different object category: ball, book, detergent, flower pot, hat, kettle, keyboard, knife, microwave, orange, peach, pineapple, pliers, remote control, stuffed toy, razor, and scissors. Keyframes are uniformly sampled from each sequence with a stride of 10.
As shown in Table~\ref{wild}, PointVGGT consistently outperforms all \textit{pairwise-then-global} baselines across all metrics. Among the baselines, UR\&R+PGO is the strongest competitor, achieving 58.3\%/76.0\%/88.5\% rotation recall at 3$^\circ$/5$^\circ$/10$^\circ$. In contrast, PointVGGT (Pi3) reaches 88.7\%/93.5\%/98.9\% at the same thresholds, achieves 100\% recall at 30$^\circ$/45$^\circ$, and yields only 0.01\,m/0.01\,m mean/median translation error, indicating near-perfect registration across all 17 object categories. 
PointVGGT (DA3) also performs strongly, achieving 80.8\%/93.1\% recall at 3$^\circ$/5$^\circ$ and 100\% recall at 30$^\circ$/45$^\circ$. These results demonstrate that the proposed method generalizes robustly beyond scene-level registration to object-centric settings in a fully zero-shot manner. 
More qualitative comparisons are provided in Fig.~\ref{wildvis}.


\subsubsection{Evaluation on Outdoor Benchmarks} 
\noindent\textbf{KITTI.}
KITTI~\cite{geiger2012we} is a standard large-scale outdoor benchmark where point clouds are acquired by a vehicle-mounted spinning LiDAR sensor in urban driving environments. This setting introduces a fundamentally different sensing modality: point clouds are substantially sparser and cover far larger spatial extents than indoor RGB-D data. 
For evaluation, we sample up to the first 20 keyframes from each test sequence. 
Since ZeroMatch is the only zero-shot method capable of operating on both indoor and outdoor point clouds (the remaining baselines are trained exclusively on indoor RGB-D data), we compare against ZeroMatch with the three global synchronization backends (GTSAM, SGHR, PGO).
As reported in the left half of Table~\ref{scan}, ZeroMatch+PGO achieves perfect rotation recall (100\%/100\% at 5$^\circ$/10$^\circ$) but with a substantial mean translation error of 0.99\,m. PointVGGT (Pi3) matches the perfect rotation recall with a mean/median rotation error of 0.3$^\circ$/0.2$^\circ$, and markedly improves translation accuracy, achieving 75.6\%/ 85.0\% recall at 0.5\,m/0.75\,m thresholds (vs.\ 64.9\%/73.6\% for ZeroMatch+PGO) and reducing mean translation error from 0.99\,m to 0.41\,m. These results confirm that PointVGGT transfers effectively to LiDAR-based outdoor registration, demonstrating the cross-modality generalization of our \textit{foundation-then-refinement} paradigm.

\noindent\textbf{Waymo.}
The Waymo Open Dataset~\cite{sun2020scalability} is a diverse autonomous driving benchmark featuring varied weather conditions, traffic densities, and geographic environments, providing a rigorous test of cross-domain robustness. 
For evaluation, we randomly select 15 test sequences and retain up to the first 20 keyframes from each sequence. 
 Following the same comparison protocol as KITTI, we evaluate against ZeroMatch with GTSAM, SGHR, and PGO backends.
As shown in the right half of Table~\ref{scan}, PointVGGT (VGGT) achieves the strongest overall performance, 100\%/100\% rotation recall at 5$^\circ$/10$^\circ$ and 98.7\%/99.7\% translation recall at 0.5\,m/0.75\,m, with a mean/median translation error of 0.07\,m/0.03\,m, far outperforming the best ZeroMatch baseline (mean 1.91\,m). The strong performance on both KITTI and Waymo, despite their markedly different environmental conditions and geographic settings, demonstrates the robust zero-shot generalization of PointVGGT across outdoor LiDAR scenarios, alongside its effectiveness on indoor RGB-D benchmarks.

\begin{table*}[t]
	\centering
	\caption{Ablation Studies on different components and configurations  evaluated on \textbf{ScanNet}~\cite{dai2017scannet}.
    }
	\resizebox{1\textwidth}{!}{
		\begin{tabular}{lccccccccccccc}
			\toprule[1.8pt]
			\multirow{3}{*}{\textbf{Method}}& \multicolumn{6}{c}{\textbf{Rotation Error}} & \multicolumn{6}{c}{\textbf{Translation Error} (\textit{m})} \\
			\cmidrule(lr){2-7} \cmidrule(lr){8-13} 
			 & 3$^\circ$ & 5$^\circ$ & 10$^\circ$ & 30$^\circ$ & 45$^\circ$ & Mean/Med & 0.05 & 0.1 & 0.25 & 0.5 & 0.75 & Mean/Med  \\
			\midrule[1.2pt]
			Foundation & 90.9 & 99.0 & 100.0 & 100.0 & 100.0 & 1.5/1.2 & 65.6 & 86.9 & 97.8 & 100.0 & 100.0 & 0.06/0.04 \\
			Foundation + Refine & 92.2 & 99.3 & 100.0 & 100.0 & 100.0 & 1.4/1.1 & 73.4 & 89.0 & 98.0 & 100.0 & 100.0 & 0.05/0.03 \\
            \cmidrule(lr){1-13}
             ADAM  & 91.1 & 98.9 & 100.0 & 100.0 & 100.0 & 1.4/1.1 & 70.8 & 88.7 & 97.8 & 100.0 & 100.0 & 0.05/0.03 \\
            GN w/ CG (ours) & 92.2 & 99.3 & 100.0 & 100.0 & 100.0 & 1.4/1.1 & 73.4 & 89.0 & 98.0 & 100.0 & 100.0 & 0.05/0.03 \\
            \cmidrule(lr){1-13}
            Reproject. Residual  & 90.9 & 99.0 & 100.0 & 100.0 & 100.0 & 1.5/1.2 & 68.0 & 87.1 & 97.8 & 100.0 & 100.0 & 0.05/0.03 \\
            Reproject. + Corr. Residual & 92.2 & 99.3 & 100.0 & 100.0 & 100.0 & 1.4/1.1 & 73.4 & 89.0 & 98.0 & 100.0 & 100.0 & 0.05/0.03 \\
			\bottomrule[1.8pt]
	\end{tabular}}
	\label{abla10}
\end{table*}

\subsection{Ablation Studies}
\label{sec:ablation}


\noindent\textbf{Effectiveness of the Foundation-Driven Initialization.}
We first conduct the ablation experiment to validate the contribution of foundation-driven pose initialization  \emph{Foundation}, which applies the scale-aligned foundation-model poses directly without any subsequent refinement.  
As demonstrated in Table~\ref{abla10}, the \emph{Foundation} configuration alone already delivers remarkably strong performance. On ScanNet, it achieves 99.0\%/100.0\% rotation recall at 5$^\circ$/10$^\circ$ with a mean rotation error of only 1.5$^\circ$, surpassing the best \emph{pairwise-then-global} baseline (ZeroMatch+PGO, 88.5\% at 10°) even without any geometric refinement. This provides direct empirical evidence for the central thesis of PointVGGT: visual geometry foundation models encode holistic, globally consistent geometric priors that bypass the fundamental limitations of pairwise pipelines. By jointly reasoning over all input frames, the foundation model circumvents the error accumulation in prior two-stage approaches, delivering high-quality global pose initialization at a fraction of the computational cost.

\noindent\textbf{Effectiveness of Pose Refinement.}
We next evaluate the performance gains brought by the pose refinement stage. 
As demonstrated in Table~\ref{vis10}, adding the voxelized dense correspondence construction and robust motion-only bundle adjustment (\emph{Foundation + Refine}) consistently improves performance across all criteria. 
For example, on ScanNet, the benefit of refinement is substantially more pronounced: 3$^\circ$ rotation recall improves from 90.9\% to 92.2\% (+1.3 points), and 0.05\,m translation recall rises from 65.6\% to 73.4\% (+7.8 points). This large gain indicates that inaccuracies in scale-factor estimation can degrade the initial poses, while our refinement stage effectively compensates for these errors. 

Moreover, we evaluate the efficacy of different losses: the correspondence residual and the geometric reprojection residual. As shown in the bottom block of Table~\ref{abla10}, incorporating these residuals consistently yields performance gains. Specifically, the reprojection residual elevates the recall ratio under the 0.05m translation threshold from 65.6\% to 68.0\%, with the best results achieved when both losses are combined (73.4\%). 

\noindent\textbf{Analysis of Optimization Strategies.} We further evaluate the effectiveness of the optimization strategy by comparing a standard ADAM baseline~\cite{kingma2014adam} with our GN-CG scheme (Gauss-Newton with conjugate gradient). 
Grounded in the IRLS objective (Eq.~\ref{irls}),  ADAM is configured with 200 iterations, a learning rate of 0.01, and momentum parameters $\beta=(0.9, 0.999)$. 
As shown in Table~\ref{abla10}, GN w/ PCG consistently outperforms first-order adaptive methods, elevating the $0.05$m translation success rate from $70.8\%$ to $73.4\%$. This confirms that leveraging second-order geometric information better accommodates the non-linearity of pose estimation than general-purpose gradient methods. 

Table~\ref{runtime3} shows the trade-off between accuracy and runtime across different Motion BA iteration counts. While accuracy at $0.05$m leaps from $65.6\%$ to $73.0\%$ with just $10$ iterations, the performance gains plateau beyond $40$ iterations. Since doubling the iterations to $80$ only yields a marginal $0.3\%$ rotation gain ($92.2\% \rightarrow 92.5\%$) while increasing latency by over $30\%$ ($11.14$s to $14.62$s), we adopt $40$ iterations as the default configuration to achieve a balance between precision and computational efficiency.

 \begin{figure}[t]
	\centering
	\includegraphics[width=\columnwidth]{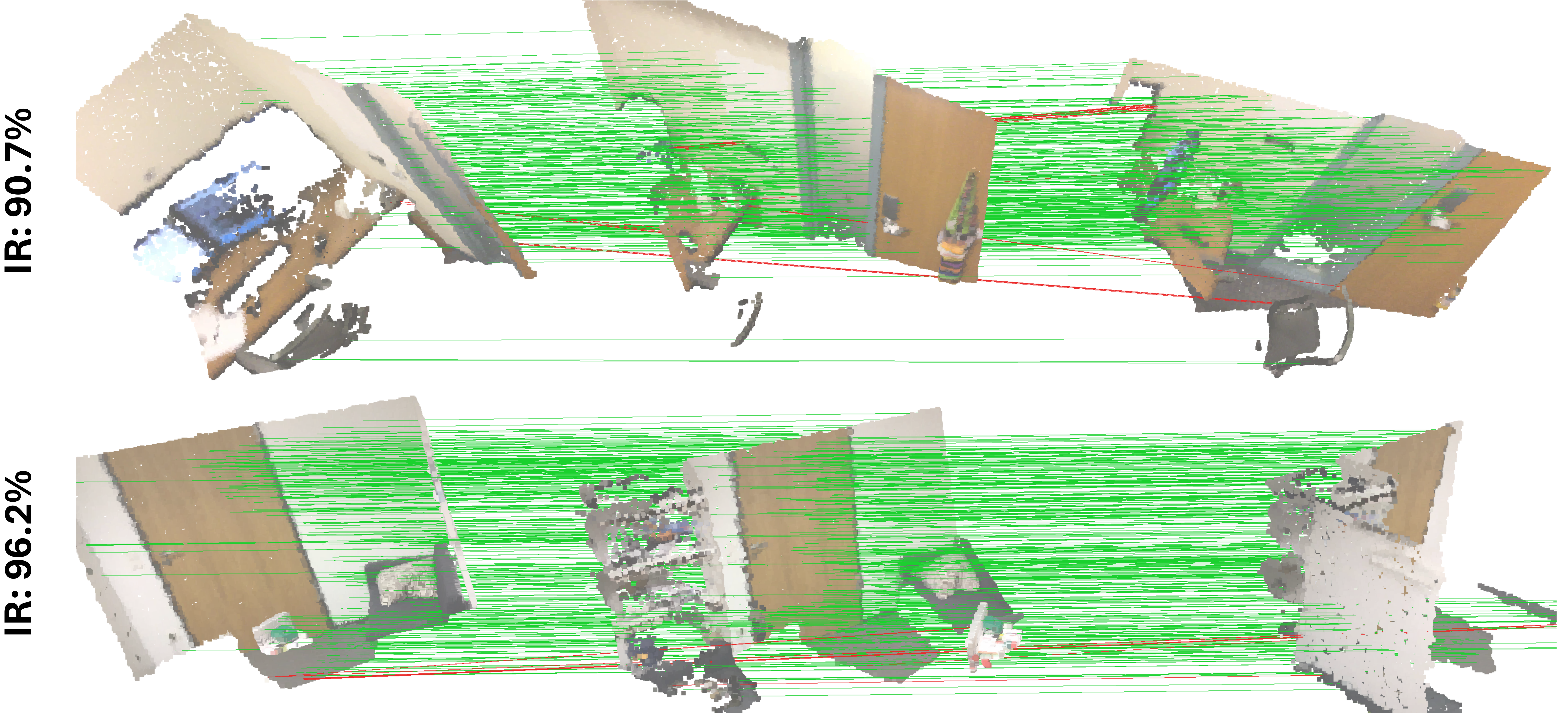}
    \vspace{-4mm}
	\caption{Visualization of cross-view correspondences produced by our voxelized spatial hashing-driven fast matching scheme. The high inlier ratios (green) across view pairs highlight the robustness and reliability of the proposed matching strategy.}
	\label{viscorr}
\end{figure}


\begin{table}[t]
	\centering
	\caption{
	Runtime (\textit{second}) of different matching schemes. \textit{NN-Match}: Nearest neighbor search; \textit{VoxelMatch}: Voxelized spatial hashing; IR: Inlier ratio.}\vspace{-1mm}
	\resizebox{1\columnwidth}{!}{
		\begin{tabular}{lccc|c}
			\toprule[1.8pt]
			 & NN-Match  & VoxelMatch &  &   \\ 
             & $\mathcal{O}(S^2N^2)$ & $\mathcal{O}(SN)$ & \textit{Reduced}$\downarrow$ & IR (\%) \\ \midrule
			ScanNet & 65.66 & 0.07 & \textcolor{teal}{65.59$\downarrow$} & 99.6   \\
            WildRGBD &  76.15 & 0.07 & \textcolor{teal}{76.08$\downarrow$} & 89.1   \\
            KITTI & 30.50 & 0.05 & \textcolor{teal}{30.45$\downarrow$} & 68.7 \\
			\bottomrule[1.8pt]
	\end{tabular}}
	\label{match}
\end{table}


\begin{table}[t]
	\centering
	\caption{
	{Runtime (\textit{second}) of each component. \textit{FoundInit}: foundation-driven pose initialization; \textit{VoxelMatch}: voxelized spatial hashing;  \textit{MotionBA}: motion-only bundle adjustment.}}\vspace{-1mm}
	\resizebox{1\columnwidth}{!}{
		\begin{tabular}{lccc|c}
			\toprule[1.8pt]
			 & FoundInit & VoxelMatch & MotionBA & Total \\ \midrule
			ScanNet & 3.27 & 0.07 & 7.80 & 11.14   \\
            WildRGBD &  3.39 & 0.07 & 0.94 & 4.40   \\
            KITTI &  1.93 &  0.05 & 10.92 & 12.90  \\
			\bottomrule[1.8pt]
	\end{tabular}}
	\label{runtime2}
\end{table}

\begin{table}[t]
	\centering
	\caption{
	{Performance and runtime under different numbers of Motion BA iterations. ($\star$) denotes the default setting.}}\vspace{-1mm}
	\resizebox{1\columnwidth}{!}{
		\begin{tabular}{lcccc|c}
			\toprule[1.8pt]
			 & 3$^\circ$@R & 5$^\circ$@R &0.05@t & 0.1@t & Time (s) \\ \midrule
			\#iter=0 & 90.9 & 99.0 & 65.6 & 86.9 & 3.33  \\
            \#iter=10 & 91.8 & 99.1 & 73.0 & 87.7 & 5.44  \\
			\#iter=20 & 91.8 & 99.2 & \textbf{73.5} & 87.9 & 7.58  \\
            \#iter=40$^\star$ & 92.2 & 99.3 & 73.4 & 89.0 & 11.14  \\
			\#iter=80 & \textbf{92.5} & \textbf{99.6} & 73.2 & \textbf{89.5} & 14.62 \\
			\bottomrule[1.8pt]
	\end{tabular}}
	\label{runtime3}
\end{table}

\noindent\textbf{Voxelized Spatial Hashing.}
We now analyze the effectiveness of the proposed voxelized spatial hashing scheme for fast correspondence construction from both efficiency and robustness perspectives.
With regard to efficiency, Table~\ref{match} shows that compared with conventional pairwise matching-based correspondence construction with a complexity of $\mathcal{O}(S^2N^2)$, the proposed voxel matching reduces the runtime from 65.66\,s to 0.07\,s on ScanNet, from 76.15\,s to 0.07\,s on WildRGB-D, and from 30.50\,s to 0.05\,s on KITTI, by reducing the matching complexity to $\mathcal{O}(SN)$ and employing parallelized CUDA acceleration.  
Despite this significant acceleration, our method still recovers high-quality correspondences, specifically yielding inlier ratios of 99.6\%, 89.1\% and 68.7\% on ScanNet, WildRGBD and KITTI. This strong performance primarily stems from the high-quality, globally aligned 3D geometry provided by the foundation model, which brings geometrically corresponding points from different views into a shared spatial neighborhood and thus enables reliable correspondence retrieval through local voxel search. 
Fig.~\ref{viscorr} illustrates our achieved high-quality correspondences. 

\noindent\textbf{Runtime Analysis.} 
We further report the runtime of each component in the proposed pipeline in Table~\ref{runtime2}, including voxelized spatial hashing-driven correspondence construction (\textit{VoxelMatch}), foundation-driven pose initialization (\textit{PoseInit}), and motion-only bundle adjustment (\textit{MotionBA}). 
It shows that \textit{VoxelMatch} is extremely efficient, requiring less than 0.07\,s across all datasets. 
By contrast, \textit{PoseInit} and \textit{MotionBA} dominate the runtime. Pose initialization typically takes 1.93-3.39\,s depending on the number of input scans, while motion-only bundle adjustment further requires 0.94-10.92\,s. In particular, Motion BA becomes the main computational bottleneck in most cases, due to iterative optimization over numerous correspondences. 
Notably, as shown in Fig.~\ref{fig:teaser}, our paradigm achieves a substantial speedup over prior \emph{pairwise-then-global} methods, reducing the runtime from minutes to mere seconds.

\section{Conclusion}
\label{sec:conclusion} 
We have introduced PointVGGT, a novel multiview RGB-D registration framework built upon a foundation-then-refinement paradigm. Unlike conventional methods, PointVGGT treats visual geometry priors as primary cues to directly recover metrically grounded global poses, eliminating the need for costly pairwise matching and global synchronization. By integrating voxelized spatial hashing for near-linear time correspondence building and a robust motion-only bundle adjustment, our method achieves outstanding zero-shot performance across six benchmarks, reducing inference time from minutes to mere seconds. Our framework is inherently plug-and-play and naturally benefits from the continuous evolution of 3D foundation models, offering a scalable and efficient solution for robust multiview alignment. 

\section*{Acknowledgments}
This research is supported by the MOE AcRF Tier 1 Grant of Singapore (RG107/25), by the RIE2025 Industry Alignment Fund – Industry Collaboration Projects (IAF-ICP) (Award I2301E0026), administered by A*STAR, as well as supported by Alibaba Group and NTU Singapore through Alibaba-NTU Global e-Sustainability CorpLab (ANGEL).

\bibliographystyle{spbasic}
\bibliography{sn-bibliography}

\end{document}